\documentclass[sageh,times,shortAfour]{sagej}
\usepackage{moreverb,url}

\usepackage[colorlinks,bookmarksopen,bookmarksnumbered,citecolor=red,urlcolor=red]{hyperref}
\usepackage[table]{xcolor}
\usepackage{tabularx}
\usepackage{booktabs, multirow, graphicx}
\usepackage{mdframed}
\usepackage{to-be-determined}

\newcommand\BibTeX{{\rmfamily B\kern-.05em \textsc{i\kern-.025em b}\kern-.08em
T\kern-.1667em\lower.7ex\hbox{E}\kern-.125emX}}

\def\volumeyear{20xx}

\begin{document}

\runninghead{M. Heyrman, C. Li, V. Klemm, D. Kang, S. Coros, and M. Hutter}

\title{Multi-Domain Motion Embedding: Expressive Real-Time Mimicry for Legged Robots}

\author{Matthias Heyrman\affilnum{1}, Chenhao Li\affilnum{1}, Victor Klemm\affilnum{1}, Dongho Kang\affilnum{2}, Stelian Coros\affilnum{2}, Marco Hutter\affilnum{1}}

\affiliation{
\affilnum{1} Robotic Systems Lab, ETH Zurich, Zurich, Switzerland
\affilnum{2} Computational Robotics Lab, ETH Zurich, Zurich, Switzerland
}

\newmdenv[
  topline=false, bottomline=false, rightline=false, leftline=true,
  linecolor=blue, linewidth=1.5pt,
  innerleftmargin=6pt, innerrightmargin=0pt,
  innertopmargin=0pt, innerbottommargin=0pt,
  skipabove=0.5\baselineskip, skipbelow=0.5\baselineskip
]{revblock}

\begin{abstract}
Effective motion representation is crucial for enabling robots to imitate expressive behaviors in real time, yet existing motion controllers often ignore inherent patterns in motion.
Previous efforts in representation learning do not attempt to jointly capture human and animal movements through structured periodic patterns and variational aperiodic descriptions.
To address this, we present Multi-Domain Motion Embedding (MDME), a motion representation that unifies the complementary embedding of structured and unstructured features using a wavelet-based encoder and a probabilistic embedding in parallel.
This produces a rich representation of reference motions from a minimal input set that generalizes across diverse motion styles.
We evaluate MDME on retargeting-free motion imitation at deployment by conditioning robot control policies on the learned embeddings to reconstruct ideal retargeted states on the robot, demonstrating accurate reproduction of long-horizon trajectories on both humanoid and quadruped platforms.
Our comparative studies confirm that MDME outperforms prior approaches in motion reproduction and generalization to unseen motions.
Furthermore, we demonstrate real-time zero-shot deployment on unseen motions, removing per-motion tuning and online retargeting.
These results show that MDME provides a generalizable and structure-aware foundation for scalable real-time robot imitation.
\end{abstract}

\maketitle

\section{Introduction}
\label{sec:introduction}
Motion imitation is an effective approach that enables robots to learn natural and expressive movement skills from human and animal demonstrations.
Behaviors like stylized walking and whole-body gestures are difficult to teach without examples to refer to.
Advances in reinforcement learning (RL)-based approaches \citep{heess2017emergence, deeprl, deepmimic, PAE, rl_sim_2, AMP, decision_transformer}, coupled with the availability of large motion capture datasets, have demonstrated the efficacy of using RL in the successful reproduction of dynamic and natural real-world motion skills and whole body teleoperation in legged robots \citep{peng2020learning, VMP, GMT, liao2025beyondmimicmotiontrackingversatile}.

Despite this success, previous motion imitation methods often struggle to adapt to unseen motion patterns for real-time mimicry, hindering their use in more general applications.
The lack of generalizability arises because they typically fail to effectively represent the inherent structures in human and animal motions and are overly specialized for tracking target motions.

Although embeddings without explicit structures have proven effective for motion reproduction \citep{VMP, li2025amo}, understanding and generalizing to novel motions requires capturing and exploiting the temporal structures inherent in natural behaviors.
Recent works highlight the importance of biasing learned representations toward recurring patterns in motion trajectories \citep{PAE, FLD, DFM, MPR}, yet the diversity of natural movement, consisting of global periodicity, temporally isolated oscillations, and non-repetitive transitions, demands embeddings that exploit these regularities while remaining flexible to irregular dynamics.

This problem is further compounded by the morphological gap.
Typically, motion imitation approaches rely on explicit retargeting to establish pose correspondences in real-time \citep{HumanoidTeleop, Dongho_2, grandia2023doc, yoon2025spatio}.
This process can obscure motion nuances and demands actor-robot specific human engineering.
Moreover, this retargeting step often hampers real-time execution, as it adds a significant computational burden.
Methods that retarget offline avoid this run-time cost when the motion set is fixed and known in advance \citep{H2O}.
However, admitting novel motions at deployment, as in real-time teleoperation, requires online retargeting, which reintroduces this computational burden.
Independently of when it is performed, retargeting commits to a fixed source-to-robot correspondence that can obscure motion nuance and produce poses that are infeasible on the target hardware.

\begin{figure}
    \centering
    \includegraphics[width=\linewidth]{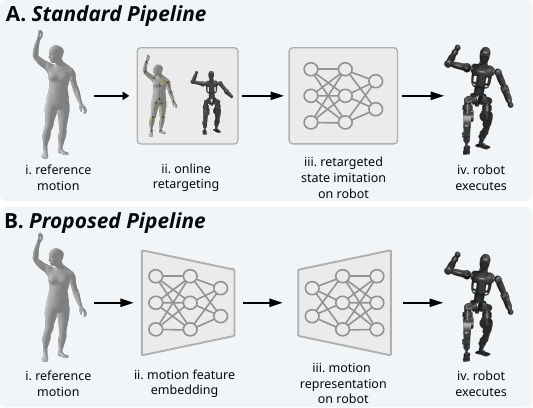}
    \caption{\textbf{A.} Standard deployment pipeline used by previous methods using a handcrafted retargeting of reference motions to the target robot's morphology and reproducing the resulting joint angles. \textbf{B.} Our proposed deployment pipeline where the reference motion is embedded and an action representation of the input motion on the robot is learned.}
    \label{fig:intro_graphic}
\end{figure}

To address these challenges, we propose Multi-Domain Motion Embedding (MDME), a framework that enables generalized real-time motion imitation.
Our mechanism employs a novel dual-encoding architecture to capture both periodic and non-repetitive components of motions.
MDME combines \textbf{(i)} a periodic encoding component that decomposes motion into frequency-domain representations across multiple temporal scales with \textbf{(ii)} a classical stochastic embedding to capture non-periodic variations and transitions.
By jointly encoding these complementary features, MDME produces a comprehensive representation that improves generalization to unseen motions.
With this richer representation, we condition policies directly on raw reference motions, shown in Figure~\ref{fig:intro_graphic}, allowing the embedding to implicitly bridge morphological differences through reinforcement learning.

Our experiments demonstrate that MDME enables robots to mimic diverse human and animal motions in real time, with stronger generalization than recent motion tracking and retargeting-based approaches.
We train policies on three robotic platforms, two humanoids and a quadruped, and observe improved tracking performance compared to previously proposed motion-reproduction architectures across benchmarks.
To assess how well these capabilities transfer to the real world, we deploy trained policies on hardware in a zero-shot manner.
The policies successfully track human motions on the Fourier N1 humanoid and dog motions on the ANYmal D quadruped.
We also find that our retargeting-free pipeline enables robust mimicry of non-expert actor motions on the ANYmal D, highlighting its practicality for real data.

Together, these results suggest that our approach expands the applicability of real-world motions to teach robots to reproduce natural behaviors.
In particular, MDME enables direct use of whole-body demonstrations from real-world actors for real-time teleoperation of legged robots, enabling them to reproduce complex, naturalistic skills without explicit motion retargeting.

In summary, our contributions are:
\begin{enumerate}
    \item A structure-aware motion embedding that captures multi-scale periodic and varying aperiodic features of motion through a novel dual-encoding architecture.
    \item A retargeting-free imitation reinforcement learning framework that directly conditions robot control on raw motion data across morphologies.
    \item Empirical validation in both simulation and real hardware demonstrating zero-shot transfer and improved generalization to novel motions across multiple robot morphologies.
\end{enumerate}
\section{Related Work}
\label{sec:related}

\subsection{Motion Imitation via RL}

Learning motor skills from expert demonstrations offers a direct alternative to reward engineering or curriculum design to acquire natural and complex behaviors in robots.
In this domain, RL is increasingly popular to develop a controller which replicates source motions collected from the real-world. 

The DeepMimic framework \citep{deepmimic} introduced deep reinforcement learning to train physics-based agents that imitate reference actors from motion-capture data, establishing a foundation for data-driven character control.
Building on this approach, subsequent work applied DeepMimic to real-world legged robots, enabling them to efficiently reproduce desired locomotion behaviors captured in the motion data \citep{peng2020learning, kang2023rl, grandia2024design, yoon2025spatio}.

More recently, this approach has been extended to address emerging challenges in humanoid robots and teleoperation settings, where policies learn to reproduce human demonstrations through RL-based imitation and motion retargeting \citep{OmniH2O, HumanoidTeleop, GMT}.
Beyond single-agent motion reproduction, recent approaches have explored more complex interaction scenarios such as physically realistic human-object manipulation and multi-contact behaviors by integrating imitation learning with object-centric or scene-aware dynamics modeling \citep{physhoi, Shadowing, li2025amo, ze2025twist}.
While direct trajectory tracking can achieve precise reproduction of motions, it typically fails to generalize beyond the specific demonstrations used during training, necessitating more robust motion encoding strategies that capture the essential characteristics of natural movement patterns.

In another line of research, generative adversarial network (GAN)-based approaches, which learn the distribution of expert state-action pairs of a motion dataset \citep{AMP, LearningAgileSkills, VersatileSkillControl, HumanMimic}, gained attention because they generalize the repertoire of motion skills beyond prerecorded motion trajectories.
Their generative capability enables style-consistent behavior while generalizing to achieve user-specified tasks.

However, in the context of puppetry or human-to-robot teleoperation, they often fall short in terms of precise motion reproduction.

\subsection{Motion Retargeting}
\label{rel:ssec:retarget}

Motion retargeting is the process of adapting a reference trajectory to a robot’s morphology to ensure compatibility.
In the context of RL-based motion imitation, a retargeting method that properly addresses both kinematic and dynamic constraints can significantly reduce the learning burden for motion reproduction.

Simple motion retargeting methods often focus solely on transferring kinematic features, such as footstep timings \citep{Dongho_retarget, Dongho_2} or limb vectors \citep{choi2019towards}, along with body pose trajectories.
However, these methods overlook the physical properties of the robotic system, such as its dynamics and actuation limits. Consequently, they often result in physically infeasible retargeted motions.

Alternatively, model-based optimal control frameworks can be adopted for motion retargeting \citep{grandia2023doc, yoon2025spatio, Dongho_SteerableImitation}, as they allow the system's dynamics and physical constraints to be explicitly considered. These methods tailor the source motion to the robot's physical limits, such as its dynamics and actuation power, which makes it significantly easier to replicate the retargeted motion on the physical hardware.
However, while these approaches are effective at generating robot-compatible trajectories that streamline downstream motion imitation tasks, their high computational cost often hampers online motion retargeting in real-time mimicry scenarios.

Some works have demonstrated that retargeting can be effectively treated as a learning problem.
In the field of computer graphics, \citet{ACE} and \citet{Pose2Motion} demonstrated adversarial training to learn pose correspondences between human actors and non-human characters.
\citet{PhysicsRLRetargeting} further trained a policy to transfer sparse human position information to physics-aware animated characters in simulation.
In robotics, \citet{DFM} used structured encoders to distill crafted reference trajectories into task-level commands for a motion policy to produce expressive locomotion.

While retargeting improves the feasibility of transferred motions, it inherently introduces a preprocessing layer that can obscure motion details and limit the system's ability to adapt to novel morphologies or unexpected patterns.
This limitation motivates learning embeddings that enable robots to interpret raw reference trajectories directly at run-time.

\subsection{Motion Embedding}
\label{rel:ssec:embedding}

The challenge of transferring biological motions to robotic platforms has driven the development of increasingly sophisticated representation learning methods.
Learning latent motion embeddings within reinforcement learning frameworks enables compact representations that capture emergent structures of natural locomotion.
These representations exploit the inherent regularities of coordinated motion to improve the fidelity and generalizability of learned control behaviors.
By tailoring the embedding design, different motion characteristics can be emphasized, allowing the policy to reproduce dynamically consistent and lifelike movements.

\subsubsection{Unstructured Motion Embeddings}
\label{rel:sssec:unstructured}

Autoencoder-based architectures have been extensively applied to motion data, using an encoder to learn a latent representation, typically with reduced dimensionality compared to the original input, and a decoder to reconstruct the original sequence.
\citet{dataset_dog} took a deterministic approach with Mode-Adaptive Neural Networks, where a gating network blends the weights of expert networks to specialize across motion modes, though the blending coefficients must be conditioned on an externally supplied phase.

Variational Autoencoders (VAEs) \citep{VAE} have become a dominant paradigm, encoding motion inputs into continuous probabilistic distributions that can be decoded to reconstruct or generate new trajectories. 
\citet{MP-VAE} demonstrated the effectiveness of hierarchical VAE-based representations for humanoid motion synthesis.
\citet{VMP} advanced hierarchical methods for reconstructing input motions with Versatile Motion Priors (VMP).
Works incorporating progressive networks \citep{GMT} and knowledge distillation \citep{PHC, UHC} have further improved generalization capabilities and mitigated catastrophic forgetting when learning from diverse datasets.

Discrete embedding approaches offer an alternative to continuous probabilistic representations.
Vector Quantized VAEs \citep{VQ-VAE-character, Lifelike} discretize the motion into a token codebook, which can be sampled by higher-level policies to generate realistic movements.
Transformer-based sequence models have also been applied to motion representation \citep{decision_transformer, Shadowing}, leveraging attention mechanisms to capture long-term dependencies. 
However, these methods require extensive datasets and are difficult to train effectively.

\subsubsection{Structured Temporal Representations}
\label{rel:sssec:temporal}

Standard autoencoder frameworks do not explicitly model the inherent temporal structures present in natural motion.
Recurrent architectures improve temporal prediction capabilities \citep{Motion-VAE} but still struggle to capture the periodic nature of locomotion patterns.
This has motivated research into frequency-domain representations that explicitly encode cyclic behavior.

\citet{PAE} introduced the Periodic Autoencoder (PAE), which uses a Fast Fourier Transform (FFT) to capture repeating motion patterns, generating a latent phase-manifold that is sampled to reconstruct motions using an RL-trained policy. 
Each latent channel is parameterized as a sinusoid whose amplitude, frequency, and offset are extracted by a differentiable FFT layer, while the phase is regressed separately, producing a periodicity-aligned manifold learned without phase supervision.
\citet{FLD} extended this periodic approach to propagate the latent embedding over time to predict future state embeddings and to represent the underlying dynamics of the original motion, unrolling the phase at constant frequency, amplitude, and offset and supervising the resulting multi-step reconstructions so that the latent parameters describe a motion's evolution rather than an isolated segment.
\citet{zhaobilevel} further developed this line of work by co-optimizing the decoder with the robot policy, adapting the reference motions toward physically consistent targets while a latent-space constraint preserves the original motion patterns.
These FFT-based frequency-domain methods excel in capturing global periodic patterns but may overlook temporal variations.

The discrete wavelet transform offers a promising alternative for temporal motion analysis, providing temporally resolved frequency decomposition that captures both sustained periodic structure and short-lived variations.
Although wavelets have been successfully applied to motion analysis in other domains, such as sign language recognition \citep{wavelet-chinese-sl}, their potential for robotic motion imitation remains largely unexplored.

Human and animal motions combine whole-body periodic sequences with local oscillations (e.g. walking looking side to side) alongside abrupt, unstructured events such as direction changes or reactive behaviors.
Existing approaches have not systematically captured this full spectrum of global, local, and non-periodic structures.
Bridging this gap requires a representation that can adaptively exploit periodic regularities while remaining flexible to non-repetitive variations, enabling generalization to novel motions.
\section{Preliminaries}
\label{pre:sec:prelim}

\subsection{Problem Formulation}
\label{pre:ssec:probform}
We formulate the motion mimicry policy as a Markov Decision Process $\mathcal{M} = \langle\mathcal{S}, \mathcal{A}, \mathcal{T}, \mathcal{R}, \gamma \rangle$.
At time $t$, the policy $\pi(\mathbf{a}_t|\mathbf{s}_t)$ outputs an action $\mathbf{a}_t \in \mathcal{A}$, defined as a low-level robot joint target command, given the current state $\mathbf{s}_t \in \mathcal{S}$ and transition probability $\mathcal{T}$.
The state $\mathbf{s}_t = (\mathbf{s}_t^r, \mathbf{s}^g_t)$ is composed of proprioceptive observations of the robot's state $\mathbf{s}_t^r$ and a window of $H$ non-retargeted target frames $\mathbf{s}^g_t = (s^g_{raw,t-H+1}, \dots, s^g_{raw,t}) {}\in \mathbb{R}^{n_g \times H}$ from the expert demonstration, where each frame $s^g_{raw,\tau} \in \mathbb{R}^{n_g}$ is the raw reference defined in Section~\ref{pre:ssec:ref}.
The objective is to maximize the expected cumulative discounted reward 
\[
\mathbb{E}_{\pi} \left[\sum_{t \geq 0}\gamma^t\mathcal{R}(\mathbf{s}_t, \mathbf{a}_t)\right], \quad \gamma\in [0, 1].
\]

The reward function $\mathcal{R}(\mathbf{s}_t,\mathbf{a}_t)$ is constructed to minimize the discrepancy between the robot’s executed motion and a precomputed ideal retargeting of the reference actor’s kinematic and dynamic states.
This ensures that the learned behavior remains both faithful to the original demonstration and feasible given the robot’s physical constraints.

\subsection{Reference Motion Datasets}
\label{pre:ssec:ref}
The motion reproduction policies are conditioned on actors' reference pose trajectories parameterized to capture relevant motion information.
In this work, we use two motion datasets: one collected from human actors to train humanoid imitation, and one collected from a dog for quadruped imitation.

For human motion, we use a subset of the AMASS dataset \citep{dataset_AMASS} parametrized using the SMPL body model \citep{SMPL-X2019}.
The reference input at time $t$ consists of joint orientations in axis-angle form, the projected gravity vector, the base linear velocity, and the base angular velocity:
\begin{equation}
    \label{eq:human_input}
    s_{raw,t}^g = [q_{aa,t},  g_t, v_t, \omega_t].
\end{equation}
Using the SMPL joint information provides a human-morphology-agnostic input set, enabling mimicry of any humanoid.
The actor's body size nevertheless affects how accurately a motion is imitated on a fixed-size robot; Appendix~\ref{app:ssec:body_size} analyzes this dependence.
Using the method described by \citet{ze2025twist, joao2025gmr}, the SMPL parameterized actor motions are retargeted onto the destination humanoid platform and infeasible trajectories are manually filtered out.
This provides reference trajectories used for the reward calculation during policy training.

For quadruped motion, we use sequences from \citet{dataset_dog}.
Given that all recordings originate from a single animal, we adopt a sparse input representation that captures foot placement and body pose information
\begin{equation}
    \label{eq:dog_input}
    s_{raw,t}^g = [f_t, g_t, z_t],
\end{equation} 
where $f_t$ are the foot positions in base frame, $g_t$ is the projected gravity vector, and $z_t$ is the base height.
The retargeting procedure tailored to quadrupeds by \citet{Dongho_SteerableImitation} provides the ideal targets for the reward and discards motions that would necessitate non-existent degrees of freedom (e.g., pronounced spinal flexion).

The datasets are divided into a training and validation set by randomly selecting 80\% of the reference motion clips as the training set and using the remaining 20\% as the validation set.
All selection is performed after filtering out dynamically or kinematically infeasible motions.
As the AMASS dataset consists of multiple subsets recorded by different groups with different capture systems, actors, and styles, we split each subset separately so that the validation set contains motions from all subsets.

The retargeted motions are never provided as a policy input and therefore do not need to be calculated at run-time.
The training signal encourages the policy to reinterpret the raw reference motions through the proposed encoding to produce dynamically consistent actions on the robot, while the ideal retargeting serves purely as a supervisory target for the reward and as a feasibility filter.

\begin{figure*}
\centering
\includegraphics[width=\textwidth]{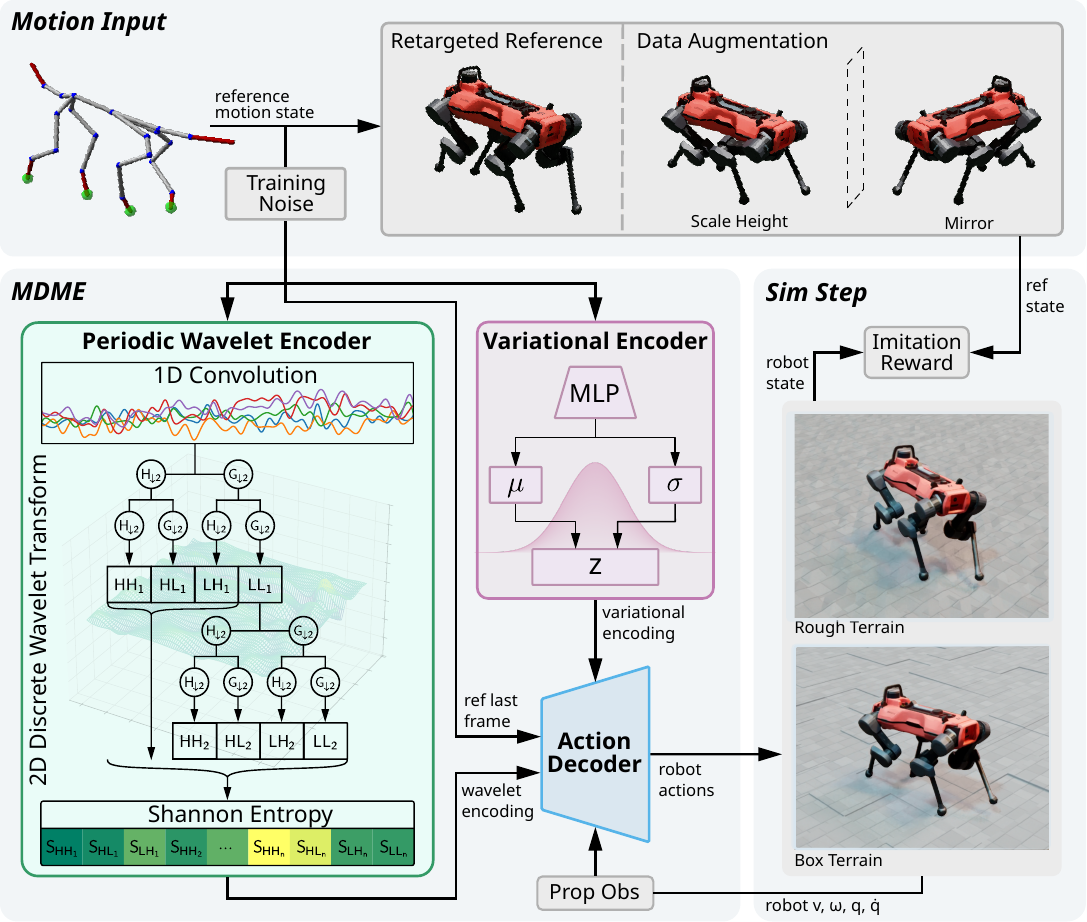}
\caption{Proposed Multi-Domain Motion Embedding training method and architecture. Produce an ideal retargeting of an input reference motion. Input the reference to the MDME architecture to teach it to reconstruct the retargeted motion on robot. Train with noisy inputs and terrain variation for additional robustness to deploy the resulting policy zero-shot on hardware.}
\label{met:fig:method_main}
\end{figure*}

\section{Multi-Domain Motion Embedding}
\label{sec:mdme}

Motion sequences captured from real-world subjects commonly exhibit repeating patterns distinctive to their movement alongside sudden discontinuous actions.
Periodic encoding methods have been shown to be powerful at representing the naturally oscillating motions of animals on quadrupeds in the frequency domain \citep{PAE, FLD}, while aperiodic representations using self-referential transformers \citep{Shadowing} or VAEs \citep{VMP} can capture unstructured motions.

Our \textbf{Multi-Domain Motion Embedding} (MDME) is an encoding framework designed to embed a motion trajectory's unstructured and structured features with high fidelity.
This enables generalizable reproduction of reference motions in real-time directly from demonstrations.

The policy $\pi$ described in Section~\ref{pre:ssec:probform} decomposes into three components.
Shown in Figure~\ref{met:fig:method_main}, the encoder portion combines a variational autoencoder that captures unstructured information and  a wavelet-based encoder that extracts periodic structures from an input motion sequence.
Both encoders receive the full history buffer $H$ of non-retargeted target states $\textbf{s}_t^g$ as inputs and each output a separate latent embedding.
The resulting embeddings are fused with the latest goal state input and the latest frame of the robot's proprioceptive state and decoded into robot actions to mimic the source motion as interpreted for the robot's morphology.
The subsequent sections describe each component of the MDME in detail.
Additionally, the code used to train and validate the MDME architecture on all discussed platforms will be released upon publication.

\subsection{Unstructured Variational Encoder}
\label{met:ssec:aperiodic}
The proposed architecture uses the encoder network of a VAE, inspired by the success of Versatile Motion Priors (VMP) proposed by \citet{VMP}, to produce a probabilistic embedding
\begin{equation*}
    \mathbf{z}^v_t = \mathbf{Enc}_{v}(\mathbf{s}^g_t).
\end{equation*}
The input buffer $\mathbf{s}^g_t$ is passed through a multilayer perceptron encoder to extract relevant features described by a learned Gaussian distribution $\mathcal{N}(\boldsymbol{\mu}_t, \boldsymbol{\sigma}_t)$ over the latent space.

From this distribution, a latent vector $\mathbf{z}^v_t \in \mathbb{R}^{n_a}$ of platform-dependent dimension $n_a$, $32$ for the quadruped and $64$ for the humanoids (Table~\ref{exp:table:hyperparams}), is sampled using the reparameterization trick \citep{VAE_rep1, VAE_rep2}
\begin{equation*}
    \mathbf{z}^v_{t} = \boldsymbol{\mu}_t + \boldsymbol{\epsilon}_t \odot \boldsymbol{\sigma}_t, \quad \boldsymbol{\epsilon}_t \sim \mathcal{N}(0, I).
\end{equation*}
Given an input of motion data in the time domain, the probabilistic latent representation $\mathbf{z}^{v}_t$ models nonrepetitive variations in the reference.
In the presented application, this includes nuances in individual limb motion or non-oscillating changes in body orientation.

\subsection{Structured Wavelet Encoder}
\label{met:ssec:periodic}
To model the oscillatory structures present in natural motion, the MDME employs a discrete wavelet transform (DWT)-based encoder.
The DWT offers a representation that captures how the frequency content of a signal evolves over time, providing a time-frequency decomposition of the input motion sequence \citep{wavelet_friendly}.
Unlike Fast Fourier Transform based approaches, which assume global periodicity, the DWT decomposes a signal using wavelets that capture how frequency content changes over time at different scales, isolating transient features, discontinuities, and short-lived structures while still capturing longer-range patterns at coarser scales.
As a result, the DWT provides a more detailed characterization of complex, quasi-periodic signals such as motion data.

To prepare the input for the DWT, the input buffer $\mathbf{s}_t^g$ is flattened and preprocessed by three 1D convolution layers $l\in \{1, 2, 3\}$
\begin{align*}
    \mathbf{y}_l = b_l + \sum^{M_{l-1}-1}_{m=0}\mathbf{K}_m \mathbin{\ast} x_{l}^{(m)},
\end{align*}
where $\mathbf{K}_m$ is the $m$-th convolution kernel, $M_0 = n_g$ is the number of input channels, $x_l = \mathbf{y}_{l-1}$, $x_l^{(m)}$ its $m$-th channel, and $x_1 = \mathbf{s}_t^g$.
These reduce the channel dimension to $C_z$ embedded phase channels, followed by intermediate batch normalization layers \citep{PAE, FLD, DFM, MPR} and ELU activation functions.
This produces a latent embedding $\mathbf{y}_t\in \mathbb{R}^{C_z \times H}$ that suppresses noise and isolates the relevant components of the input signal.

The preprocessed signal is decomposed using a differentiable 2D DWT \citep{pytorch_wavelet} using a \texttt{db2} mother wavelet \citep{db_wavelet} across $J$ levels
\begin{equation*}
    \mathbf{W}^{J}_t = \text{DWT}_{\texttt{db2}}^{J}(\mathbf{y}_t).
\end{equation*}
The mother wavelet is a prototype function from which the low-pass and high-pass filters are derived.
The wavelet's support width determines its temporal localization, its vanishing moments determine the polynomial order it ignores, and its symmetry determines the phase behavior of the filters.

From experimental evidence, the differing motion styles of quadrupedal and bipedal locomotion favor different mother wavelets.
We primarily adopt \texttt{db2} for its balanced performance across both.
The \texttt{db2} wavelet is an orthogonal wavelet with 2 vanishing moments and a compact support of width three from the Daubechies wavelet family.
The wavelet properties and the selection ablation are detailed in Appendix~\ref{ssec:mother_wavelet_choice}.
The DWT decomposes the signal by convolving it with this filter pair ($\mathbf{h}$, $\mathbf{g}$) and subsampling the result by two.
Applied separably along both axes, this produces the low-frequency approximation subband $LL$ and the high-frequency detail subbands $LH$, $HL$, and $HH$ \citep{wavelet_friendly, wavelet_really_friendly}
\begin{align*}
\begin{split}
    \text{Approximation:} \quad & LL_l=(LL_{l-1} \ast \mathbf{h}) \downarrow_2 \ast \mathbf{h}^{\top} \downarrow_2\\
    \text{Detail:} \quad & LH_l = (LL_{l-1} \ast \mathbf{h})\downarrow_2 \ast \mathbf{g}^{\top} \downarrow_2 \\
    & HL_l = (LL_{l-1} \ast \mathbf{g}) \downarrow_2 \ast \mathbf{h}^{\top} \downarrow_2\\
    & HH_l =(LL_{l-1} \ast \mathbf{g}) \downarrow_2 \ast \mathbf{g}^{\top} \downarrow_2,
\end{split}
\end{align*}
where $\downarrow_2$ denotes dyadic downsampling, $(\cdot)^{\top}$ the filter transpose, the recursion is initialized with $LL_0 = \mathbf{y}_t$, and each approximation is decomposed in turn for $l = 1,\dots,J$.
Collecting the coarsest approximation and all detail subbands gives the DWT output
\begin{align*}
    \mathbf{W}^{J}_t = \{LL_J, \{LH_l,HL_l,HH_l\}_{l=1}^J\},
\end{align*}
where $J$ is the number of decomposition levels, $LL_{J}$ captures the coarsest scale, and $\{LH_l, HL_l, HH_l \}$ contain the finer detail components at level $l$.

\subsubsection{Wavelet Entropy}
\label{mdme:sssec:wavelet_entropy}
To manage the high dimensionality of the DWT output, a single wavelet entropy is calculated for each subband \citep{wavelet-chinese-sl}, compressing each into a compact fixed-size descriptor.
This creates an efficient statistical representation of the DWT outputs with minimal loss of performance.

For a subband $b \in \mathbf{W}^J_t$ with coefficients $\{w^b_i\}$, the squared coefficients are normalized into an energy distribution $p^b_i$, from which the Shannon entropy $S^b_t$ is computed:
\begin{align*}
\begin{split}
    p^{b}_i&=\frac{(w^{b}_i)^2}{\sum_k (w^{b}_k)^2}, \\
    S^{b}_t &= -\sum_i p^{b}_i \, \log_2 p^{b}_i.
\end{split}
\end{align*}
The Shannon wavelet entropy represents the disorder of the coefficients within a subband:
a low wavelet entropy captures persistent, longer-duration oscillatory patterns, while a high entropy captures transient higher-energy events.

Stacking the per-subband entropies over the $1 + 3J$ subbands yields the wavelet encoding
\begin{equation*}
    \mathbf{z}^w_t = \mathbf{Enc}_w(\mathbf{s}^g_t) = \mathbf{S}_t {}= \big[\,S^b_t\,\big]_{b=1}^{1+3J} \in \mathbb{R}^{1+3J}.
\end{equation*}
Applied across the DWT output, this acts as a multi-resolution statistical embedding of the motion spatial-temporal-frequency structure, encoding both sustained dynamics and unstructured motions.

The entropy summary is applied to reduce the size of the structured representation rather than to strengthen it.
Forwarding the raw coefficients instead does not necessarily improve nor degrade tracking depending on the platform, and should be verified independently based on implementation details.

\subsection{Action Space Decoder}
\label{met:ssec:decoder}
The latent unstructured and structured wavelet embeddings are appended to the latest frame of the goal buffer $\mathbf{s}^g_t$ to create the complete reference input representation.
The proprioceptive observation $\mathbf{s}^p_t$ is defined as a subset of the robot state $\mathbf{s}^r_t$,
\[
\mathbf{s}^p_t = [q_{t}, \dot{q}_{t}, v_{t}, \omega_{t}, g_{t}] \in \mathbf{s}^r_t,
\]
comprising the joint positions $q_t$ and velocities $\dot{q}_t$, the base velocity terms $v_t$ and $\omega_t$, and the projected gravity $g_t$.
The action output of the previous time step $\mathbf{a}_{t-1}$ is also provided.

These are all input to an action decoder: a fully-connected MLP that outputs joint actions $\mathbf{a}_t \in \mathbb{R}^{N_q}$ for the robot to take at time step $t$
\begin{equation*}
\mathbf{a}_t=\mathbf{Dec}(\textbf{z}^w_t, \mathbf{z}^v_t, \mathbf{s}^g_t,\mathbf{s}^p_t, \mathbf{a}_{t-1}),
\end{equation*}
where the action $\mathbf{a}_t$ is the robot's goal joint positions.
The goal joint positions are provided to the low-level joint controllers.
Deployment details for the individual robots are further discussed in Section~\ref{sec:experiments}.

We name this component an action space decoder, in contrast with the trained policy $\pi$ of prior work, because the encoders and the joint-action MLP are trained as one end-to-end network.
Prior methods \citep{MP-VAE, FLD, PAE, CALM, VMP} instead train in multiple stages: an initial policy or encoder is pre-trained offline, frozen, and a separate policy is trained on its output latents.
The second stage adds training time without guaranteeing better performance, as shown in Appendix~\ref{app:ssec:stage_training}, which compares the two schemes directly for the presented problem.
\section{Training and Experiment Setup}
\label{sec:experiments}
The proposed MDME architecture is evaluated for real-time mimicry on two distinct robotic morphologies: quadrupeds (ANYmal D) and humanoids (Unitree H1 and Fourier N1).
All reference trajectories are learned concurrently until the end of training, allowing the policy to fully learn the relationship between the raw kinematic and dynamic reference states and the actions required to reproduce them on the robot, without a training curriculum.
Policies are trained using Proximal Policy Optimization (PPO) in IsaacLab \citep{mittal2023orbit} with Isaac Sim 5.0.0 using a single RTX 3090 on a remote HPC cluster running Ubuntu 24.04.5 LTS, and simulation validation experiments are performed locally on an RTX 5080.

The reward function is a weighted sum of reference tracking, regularization, and termination terms, formally provided in Appendix~\ref{app:ssec:rewards}.
Tracking terms are applied to reward the reproduction of the retargeted joint positions, foot positions, base velocities, base orientation, and height on the robot.
Penalties on the joint torques, accelerations, and action rates discourage jerky, high-effort motions, a contact penalty discourages touching the ground with non-foot links, and a strong termination penalty is applied to falls and self-collisions.

Policies are trained and executed at 50 Hz in simulation and on hardware.
Training hyperparameters are provided in Table~\ref{exp:table:hyperparams}.
For hardware deployment, the policy commands a joint-level PD controller running at 500 Hz; the per-joint gains and action scales for both platforms are listed in Table~\ref{tab:pd_gains}.

\begin{table}[]
    \centering
    \small
    \begin{tabular}{p{3.1cm}ll}
        \toprule
        Parameter & ANYmal D & H1 \& N1 \\
        \midrule
        Steps per environment & 24 & 24 \\
        Maximum iterations & 30\,000 & 40\,000 \\
        Empirical normalization & False & False \\
        \midrule
        Stochastic encoder dims & [512, 256, 32] & [512, 256, 64] \\
        Periodic 1D convolution latent channels & 25 & 15 \\
        Periodic DWT levels of decomposition & 4 & 2 \\
        Action decoder dims & [512, 256, 128] & [512, 256, 128] \\
        Activation function & \texttt{elu} & \texttt{elu} \\
        History buffer size & 25 & 15 \\
        Entropy coefficient & 3e-3 & 1e-3 \\
        Learning rate & 1e-3 & 1e-4 \\
        \midrule
        Clipping parameter & \multicolumn{2}{c}{0.2} \\
        Learning epochs & \multicolumn{2}{c}{5} \\
        Mini batches & \multicolumn{2}{c}{4} \\
        Learning rate schedule & \multicolumn{2}{c}{Adaptive} \\
        Discount factor $\gamma$ & \multicolumn{2}{c}{0.99} \\
        GAE $\lambda$ & \multicolumn{2}{c}{0.95} \\
        Desired KL divergence & \multicolumn{2}{c}{0.01} \\
        \bottomrule
    \end{tabular}
    \caption{Training hyperparameters for motion mimicry on the ANYmal D, Unitree H1, and Fourier N1 platforms.}
    \label{exp:table:hyperparams}
\end{table}

\begin{table}[t]
    \centering
    \small
    \caption{Joint-level PD gains and action scales used for deployment of the Fourier N1 and ANYmal D platforms.
    Joints paired by symmetry share identical values.}
    \label{tab:pd_gains}
    \begin{tabular}{lccc}
    \toprule
    Joint & $K_p$ & $K_d$ & Action scale \\
    \midrule
    \multicolumn{4}{c}{Fourier N1 (Humanoid)} \\
    \midrule
    Hip pitch      & 180 & 10.0 & 0.5 \\
    Hip roll       & 120 & 10.0 & 0.5 \\
    Hip yaw        & 90  & 8.0  & 0.5 \\
    Knee pitch     & 120 & 8.0  & 0.5 \\
    Ankle roll     & 45  & 2.5  & 0.5 \\
    Ankle pitch    & 45  & 2.5  & 0.5 \\
    Waist yaw      & 90  & 8.0  & 0.5 \\
    Shoulder pitch & 90  & 8.0  & 0.5 \\
    Shoulder roll  & 45  & 2.5  & 0.5 \\
    Shoulder yaw   & 45  & 2.5  & 0.5 \\
    Elbow pitch    & 45  & 2.5  & 0.5 \\
    Wrist yaw      & 45  & 2.5  & 0.5 \\

    \midrule
    \multicolumn{4}{c}{ANYmal D (Quadruped)} \\
    \midrule
    Hip abduction     & 85 & 0.6 & 1.0 \\
    Hip flexion       & 85 & 0.6 & 1.0 \\
    Knee flexion      & 85 & 0.6 & 1.0 \\
    \bottomrule
    \end{tabular}
\end{table}

\subsection{Animal Mimicry}
\label{exp:sec:quad}
The ANYmal D \citep{anymal_specs} robot is used as a quadruped platform to test real-time mimicry of animal motions.
The policy receives the kinematic reference state parameters described in Equation~\ref{eq:dog_input} as input.

After filtering undesirable or infeasible motions, the dog training dataset \citep{dataset_dog} consists of approximately 10 minutes of motion.
The dataset is augmented by exploiting the robot’s bilateral symmetry through reflection across the x- and y-axes, and by scaling the height of the source animal to increase motion diversity \citep{Dongho_SteerableImitation}.
This augmentation expands the training set to approximately 52 minutes of motion data.

\begin{table}[]
    \centering
    \small
    \begin{tabular}{ll}
        \toprule
        Property & Noise \\
        \midrule
        Linear Reference Input Command & $\mathcal{U}(-0.05, 0.05)$ \\
        Angular Reference Input Command & $\mathcal{U}(-0.2, 0.2)$ \\
        Base Mass Distribution & $\mathcal{U}(-7.0, 7.0)$\\
        Random Push Interval & $\mathcal{U}(10.0, 15.0)$ \\
        Random Push X/Y Velocity & $\mathcal{U}(-0.5, 0.5)$ \\
        Observed Base Linear Velocity & $\mathcal{U}(-0.1, 0.1)$ \\
        Observed Base Angular Velocity & $\mathcal{U}(-0.2, 0.2)$ \\
        Observed Projected Gravity & $\mathcal{U}(-0.05, 0.05)$\\
        Observed Joint Position & $\mathcal{U}(-0.1, 0.1)$ \\
        Observed Joint Velocity & $\mathcal{U}(-1.5, 1.5)$ \\
        \bottomrule
    \end{tabular}
    \caption{Quadruped domain randomization parameters.}
    \label{exp:table:quad_domain_randomization}
\end{table}
\begin{table}[]
    \centering
    \small
    \begin{tabular}{ll}
        \toprule
        Property & Noise \\
        \midrule
        Joint Reference Input & $\mathcal{U}(-0.05, 0.05)$ \\
        Projected Gravity Reference Input & $\mathcal{U}(-0.05, 0.05)$ \\
        Static Friction & $\mathcal{U}(0.6, 1.0)$ \\
        Dynamic Friction & $\mathcal{U}(0.5, 1.0)$ \\
        Random Push Interval & $\mathcal{U}(20.0, 25.0)$ \\
        Random Push X Velocity & $\mathcal{U}(-1.0, 1.0)$ \\
        Random Push Y Velocity & $\mathcal{U}(-0.25, 0.25)$ \\
        \textit{Base Mass Addition} & $\mathcal{U}(-2.5, 2.5)$\\
        \textit{Base CoM X/Y Distribution} & $\mathcal{U}(-0.05, 0.05)$ \\
        \textit{Base CoM Z Distribution} & $\mathcal{U}(-0.05, 0.05)$ \\
        \textit{PD Gain $K_p$ Scale} & $\mathcal{U}(0.8, 1.2)$ \\
        \textit{PD Gain $K_d$ Scale} & $\mathcal{U}(0.5, 2.0)$ \\
        \textit{Joint Friction} & $\mathcal{U}(-0.05, 0.05)$ \\
        \bottomrule
    \end{tabular}
    \caption{Humanoid training domain randomization parameters for Unitree H1 and Fourier N1 platforms. Randomization terms introduced during the fine-tuning stage (Section~\ref{exp:ssec:human}) are shown in italics.}
    \label{exp:table:human_domain_randomization}
\end{table}

\subsection{Human Mimicry}
\label{exp:ssec:human}

The human mimicry task is conducted in simulation using the Unitree H1 humanoid robot \citep{h1_specs} and validated on hardware using the Fourier N1 \citep{fourier_n1}.
These platforms were selected because of their morphological similarities with humans, with fully articulable shoulders and hips and a torso pitch joint.
On both platforms, the ankle joints do not receive an explicit imitation reward to encourage the emergence of environment-sensitive compliance via proprioceptive feedback.

The human motion data used in this work is presumed to be biased toward right-handed and right-footed motions, reflecting the majority of the population.
To mitigate effects from this bias, the humanoid policies are trained with left-right symmetry data augmentation: each update batch is augmented with a mirrored copy of the observations and joint-target actions so the policy learns from every transition and its mirror image.

Humanoid robots require more reward tuning to achieve desirable results, so the reward function is designed to aid learning for the two tested platforms as detailed in Appendix~\ref{app:ssec:rewards}.
To bootstrap the acquisition of initial locomotion skills, reward terms for base velocity are separated, as are those for upper and lower body joint tracking.

For robust zero-shot deployment on hardware, policies are trained with domain randomization to the simulated sensor observations and physics parameters.
The randomized properties and their ranges are listed in Table~\ref{exp:table:quad_domain_randomization} for the quadruped and Table~\ref{exp:table:human_domain_randomization} for the humanoid platforms.

After a successful human imitation policy is trained with limited initial domain randomization, it is fine-tuned by introducing the additional randomization terms shown in italics in Table~\ref{exp:table:human_domain_randomization}, the PD gain, joint friction, base center-of-mass, and base mass addition, preparing the policy for deployment.

\subsection{Evaluation Methodology}
\label{exp:ssec:eval_method}
For all tested platforms, the validation motions are withheld from the training set to assess generalization performance.
Performance is summarized by a per-motion aggregate tracking score against the ideal retargeted trajectory,
\begin{equation}
    \label{exp:eq:agg_perf}
    \text{Agg.\ Perf.} = \left(\frac{1}{|\mathcal{Q}|}\sum_{q\in\mathcal{Q}} \left[1 - \frac{e_q}{\bar{r}_q}\right]_{+}\right) s,
\end{equation}
where $e_q$ is the tracking error for quantity $q\in\mathcal{Q}$ composed of joint position, base linear velocity, base angular velocity, orientation, and base height, averaged over all steps and episodes of every motion, and $[x]_+ = \max(x, 0)$.

Each term is normalized by $\bar{r}_q$, the motion's mean reference magnitude for that quantity reduced identically to $e_q$.
The quantity-wise mean is multiplied by the success rate $s$, the fraction of episodes that terminate without a failure.
Normalizing each quantity by its own reference magnitude makes the terms comparable both across the motion library and across quantities of differing units and scales, as the same error is negligible in a high-amplitude kick but severe in a subtle gesture.
The survival factor ensures failures reduce the score, as error is only accumulated while the robot survives.

The score has a unit scale where $1$ denotes perfect tracking and $0$ means the error equals or exceeds the reference magnitude on every quantity, and $0.6$ marks the episode-level tracking-success threshold on the same per-quantity accuracy scale.
Aggregating over the full set of tracked quantities credits whole-body fidelity rather than joint tracking alone, so a policy that trades a small amount of joint accuracy for better orientation or height tracking is scored fairly.
Scores are computed per motion and averaged over independently trained policies.

\section{Results}
\label{sec:results}

The trained MDME policies are evaluated for their ability to mimic previously unseen motion trajectories from the AMASS human motion database and the MANN dog motion dataset.
We measure quantitative performance with the methodology of Section~\ref{exp:ssec:eval_method}.

\begin{figure*}
    \centering
    \includegraphics[width=\linewidth]{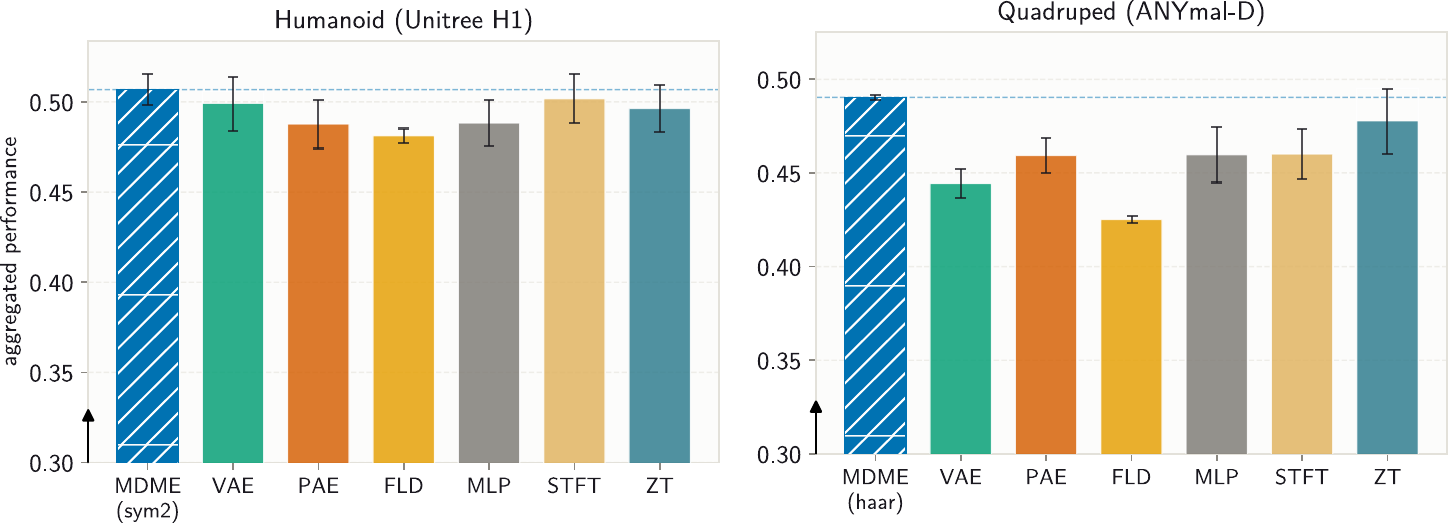}
    \caption{Method comparison on both platforms. Aggregate tracking score of Equation~\ref{exp:eq:agg_perf} on the unseen validation motion set averaged over independently trained policies (higher is better), with error bars showing the standard deviation across training seeds. MDME is marked by the dashed line. The compared encoders are described in Section~\ref{ssec:res:method_comp} and formalized in Appendix~\ref{app:ssec:encoder_comparison}.}
    \label{fig:method_comparison}
\end{figure*}

\subsection{Method Comparison}
\label{ssec:res:method_comp}
We compare MDME against a set of encoder baselines covering unstructured, structured-periodic, and transform-based approaches.
Each baseline replaces only the encoder, while the history buffer, training budget, and decoder remain fixed.
The unstructured VAE encoder of Versatile Motion Priors (VMP) \citep{VMP} embeds the input buffer into a Gaussian latent space without any structural bias.
The structured Periodic Autoencoder (PAE) \citep{PAE} reduces each latent channel to a single global sinusoid using the FFT, and Fourier Latent Dynamics (FLD) \citep{FLD} extends PAE by propagating this parametrization forward in time.
The transform-based encoders replace the DWT within the MDME architecture: the discrete Fourier transform (DFT) projects onto global sinusoids, the short-time Fourier transform (STFT) onto sinusoids within a fixed window, and the Z-Transform encoder (ZT) onto damped oscillatory modes with learnable poles.
A pure MLP policy removes the encoders entirely.
The encoder variants are covered in more detail in Appendix~\ref{app:ssec:encoder_comparison}.

The results in Figure~\ref{fig:method_comparison} show that the proposed MDME architecture achieves the highest aggregated performance for both the humanoid and quadruped motion imitation tasks over the validation set.
The strongest baseline is the Z-Transform, which overlaps the MDME mean performance within one standard deviation for both platforms, though the mean performance remains below MDME's.
The STFT and VAE also show strong performance on the humanoid, though weaker on the quadruped.
This gap may be due to the difficulty of representing the source motion given the quadruped's sparse input set.


\section{Ablation Studies}
\label{res:ssec:ablations}

Ablation studies quantify the performance contribution of each component of the MDME architecture, removing or altering one component at a time (Table~\ref{tab:h1_mdme_abl_overall}).
All ablations use the standardized \texttt{db2} basis, so each row isolates the effect of a single change.
Table~\ref{tab:h1_mdme_abl_overall} also includes an MDME policy trained on retargeted inputs, an upper bound discussed at the end of this section.


\begin{table*}[t]
    \centering
    \caption{MDME ablations on both platforms using pooled per-motion mean~$\pm$~the standard deviation of the per-seed evaluation means (Section~\ref{exp:ssec:eval_method}; MAE for joint position and height; L2-norm for velocity and orientation).
    The survival success rate is reported as the pooled mean.
    Each row removes or alters one conditioning component of the full MDME while holding the rest fixed; the MDME reference uses the standardized \texttt{db2} basis so the ablations isolate the data-composition change. \textbf{Bold}: best, \underline{underlined}: second-best, and \textit{italic}: third-best per metric within each platform.}

    \label{tab:h1_mdme_abl_overall}\label{tab:anyd_mdme_abl_overall}
    \resizebox{\textwidth}{!}{%
    \begin{tabular}{llcccccc}
      \toprule
      Platform & Method & Jnt.\ Pos.\ [rad]$\downarrow$ & Lin.\ Vel.\ [m/s]$\downarrow$ & Ang.\ Vel.\ [rad/s]$\downarrow$ & Orientation$\downarrow$ & Height [m]$\downarrow$ & Success$\uparrow$ \\
      \midrule
      \multirow{6}{*}{\rotatebox{90}{Humanoid}}
        & MDME (Ours) & \textit{0.2151$\pm$0.0067} & \textit{0.3231$\pm$0.0055} & \textbf{0.7919$\pm$0.1066} & \textit{0.1435$\pm$0.0019} & \textbf{0.0580$\pm$0.0001} & \underline{0.9959} \\
        & MDME wavelet-only & 0.2209$\pm$0.0066 & \textbf{0.3213$\pm$0.0062} & 0.8179$\pm$0.1078 & 0.1588$\pm$0.0025 & 0.0623$\pm$0.0001 & 0.9940 \\
        & MDME w/o last frame & 0.2164$\pm$0.0068 & 0.3254$\pm$0.0056 & \underline{0.8145$\pm$0.1071} & \textbf{0.1424$\pm$0.0019} & \textit{0.0602$\pm$0.0001} & \textbf{0.9981} \\
        & MDME w/o history & \textbf{0.2119$\pm$0.0007} & 0.3286$\pm$0.0024 & \textit{0.8166$\pm$0.0792} & 0.1599$\pm$0.0044 & 0.0617$\pm$0.0003 & 0.9906 \\
        & MDME w/o entropy & \underline{0.2129$\pm$0.0007} & 0.3439$\pm$0.0110 & 0.8363$\pm$0.0031 & \underline{0.1426$\pm$0.0007} & 0.0626$\pm$0.0010 & \underline{0.9959} \\
        & MDME w/ det. encoder & 0.2219$\pm$0.0067 & \underline{0.3226$\pm$0.0056} & 0.8227$\pm$0.1056 & 0.1516$\pm$0.0024 & \underline{0.0601$\pm$0.0001} & \textit{0.9955} \\
        \cmidrule{2-8}
        & MDME w/ retargeted inputs & 0.1498$\pm$0.0105 & 0.3209$\pm$0.0099 & 0.7512$\pm$0.1121 & 0.1360$\pm$0.0016 & 0.0585$\pm$0.0005 & 0.9931 \\
      \midrule
      \multirow{6}{*}{\rotatebox{90}{Quadruped}}
        & MDME (Ours) & \underline{0.1255$\pm$0.0113} & \underline{0.3022$\pm$0.0206} & \underline{0.6504$\pm$0.0120} & \underline{0.7920$\pm$0.0039} & \underline{0.0413$\pm$0.0018} & \underline{0.9724} \\
        & MDME wavelet-only & 0.1359$\pm$0.0107 & 0.3288$\pm$0.0179 & 0.6893$\pm$0.0103 & \textbf{0.7904$\pm$0.0031} & 0.0425$\pm$0.0016 & \textbf{0.9830} \\
        & MDME w/o last frame & \textit{0.1268$\pm$0.0117} & \textit{0.3041$\pm$0.0210} & \textit{0.6774$\pm$0.0136} & 0.7942$\pm$0.0039 & 0.0415$\pm$0.0017 & 0.9280 \\
        & MDME w/o history & 0.1543$\pm$0.0098 & 0.4430$\pm$0.0113 & 0.6998$\pm$0.0091 & 0.7924$\pm$0.0035 & \textit{0.0414$\pm$0.0016} & \textit{0.9694} \\
        & MDME w/o entropy & \textbf{0.0991$\pm$0.0001} & \textbf{0.2445$\pm$0.0002} & \textbf{0.5828$\pm$0.0005} & 0.7942$\pm$0.0003 & \textbf{0.0347$\pm$0.0005} & 0.9405 \\
        & MDME w/ det. encoder & 0.1328$\pm$0.0108 & 0.3070$\pm$0.0199 & 0.6858$\pm$0.0080 & \underline{0.7920$\pm$0.0023} & 0.0480$\pm$0.0020 & 0.8739 \\
        \cmidrule{2-8}
        & MDME w/ retargeted inputs & 0.1060$\pm$0.0002 & 0.2410$\pm$0.0004 & 0.4269$\pm$0.0007 & 0.7846$\pm$0.0045 & 0.0399$\pm$0.0001 & 0.9664 \\
      \bottomrule
    \end{tabular}}
\end{table*}

On the humanoid, the full MDME architecture achieves the best angular velocity and base height tracking, and where it does not lead, on joint position, linear velocity, and orientation, its gap to the leading variant stays below the standard deviation across training seeds.
On the quadruped it ranks second on every metric, behind the no-entropy variant on the four error metrics discussed below and behind the wavelet-only variant on orientation and survival.

Omitting the history buffer of previous goal states $\mathbf{s}^g_t$ has the most severe impact on the quadruped, as it inhibits the structured encoder's ability to capture frequency information.
The joint position error rises from $0.1255$ to $0.1543$~rad and the linear velocity error from $0.3022$ to $0.4430$~m/s.
On the humanoid, whose shorter buffer carries less frequency structure, the removal slightly lowers the joint position error ($0.2151$ to $0.2119$~rad) but degrades the orientation error ($0.1435$ to $0.1599$) and the survival rate ($0.9959$ to $0.9906$).
Removing the latest reference frame from the decoder input has a smaller effect.
The joint position error changes by about $1\%$ on both platforms, indicating that this input refines the immediate pose target rather than the overall motion.

Removing the unstructured embedding $\mathbf{z}^v$ shows the standalone effectiveness of the wavelet encoder $\mathbf{z}^w$ and reveals that the benefit of combining the two encoders depends on the platform.
On the quadruped, the removal is costly ($0.1225$ to $0.1359$~rad joint position and $0.3022$ to $0.3288$~m/s linear velocity), and the full MDME also outperforms the unstructured encoder alone (Figure~\ref{fig:method_comparison}), so the combination is strongest there.
On the humanoid, the wavelet-only model loses less ($0.2151$ to $0.2209$~rad).
This motivates performing the attribution analysis in Section~\ref{ssec:attribution}.

Replacing the stochastic encoder with a deterministic encoder variant using a similarly sized MLP conditions the policy on a deterministic unstructured embedding, isolating the contribution of the stochastic sampling against the otherwise identical VAE encoder.
It tracks worse than its stochastic counterpart on both platforms, indicating a consistent benefit from the sampled latent.

The final component ablation forwards the raw DWT coefficients to the decoder in place of the per-subband entropies, removing the compression while leaving the transform itself intact.
Since the decoder then receives the full multi-resolution decomposition rather than a per-subband summary, the richer structured representation is expected to track better, and on the quadruped it does so by a clear margin.
The joint position error falls from $0.1255$ to $0.0991$~rad, the linear velocity error from $0.3022$ to $0.2445$~m/s, the angular velocity error from $0.6504$ to $0.5828$~rad/s, and the base height error from $0.0413$ to $0.0347$~m, with the orientation error unchanged, although the survival rate falls from $0.9724$ to $0.9405$.
This comes at a substantial increase in size, from $0.63$M to $1.09$M parameters on the quadruped, but not at a run-time cost: the batch-1 latency in fact falls from $0.96$ to $0.61$~ms, as skipping the per-subband entropy computation outweighs the wider decoder layer it feeds (Appendix~\ref{app:ssec:compute_cost}, Table~\ref{tab:policy_cost}).

On the humanoid the two variants sit much closer, with only slightly improving the joint position and orientation errors ($0.2151$ to $0.2129$~rad and $0.1435$ to $0.1426$), degrading the velocity and base height tracking, and leaving the survival rate unchanged.
The entropy summary therefore acts as a size reduction rather than a performance mechanism, holding the structured encoding to $13$ and $7$ values in place of $919$ and $180$ raw coefficients, and whether that reduction costs accuracy depends on the platform.

\begin{figure*}
    \centering
    \includegraphics[width=\linewidth]{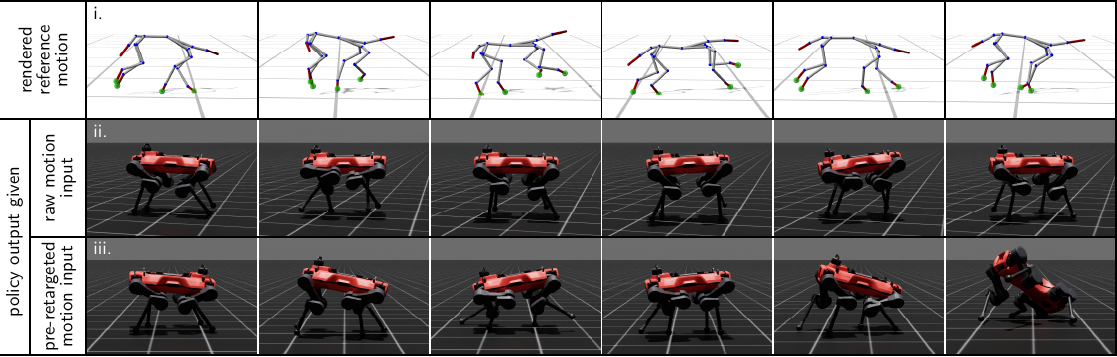}
    \caption{Simulation comparison of out-of-distribution cantering gait for retargeted and raw goal input sets: \textbf{i.} Rendering of recorded dog actor's skeleton performing reference motion. \textbf{ii.} Trained MDME policy output given raw motion input (Equation~\ref{eq:dog_input}) enabling learned retargeting. \textbf{iii.} Trained MDME policy output given retargeted joint inputs (Equation~\ref{eq:res:ret_input}).}
    \label{fig:res:ood}
\end{figure*}

\subsection{Retargeted Input Set}
\label{res:ssec:retargeted}
The final ablation compares the proposed retargeting-free pipeline with a typical retargeted input set.
For this task, we train and validate MDME imitation policies on both platforms using the same parameters and datasets as for the experiment in Section~\ref{ssec:res:method_comp}.
However, instead of inputting raw motion data (Equations~\ref{eq:human_input} and~\ref{eq:dog_input}), we provide the retargeted joint and base information used for reward computation from the retargeting method described in Section~\ref{pre:ssec:ref} as the policy input:
\begin{equation}
\label{eq:res:ret_input}
    \mathbf{s}^g_{ret,t} = [q_t, g_t, v_t, \omega_t].
\end{equation}
However, this retargeting method cannot run in real-time.

As expected, the retargeted input set outperforms the retargeting-free pipeline on in-distribution motions (Table~\ref{tab:h1_mdme_abl_overall}).
Since the goal joint states are provided directly rather than inferred from the raw motion, the retargeted policy achieves lower tracking error on nearly every metric, most markedly in joint position ($0.2151$ against $0.1498$~rad on the humanoid and $0.1255$ against $0.1060$~rad on the quadruped).
The gap is platform-dependent beyond the joint targets.
The quadruped's base velocity tracking improves substantially ($0.3022$ to $0.2410$~m/s linear and $0.6504$ to $0.4269$~rad/s angular), as the original input does not explicitly provide a velocity or angular velocity term, instead conditioning the policy on recreating accurate dynamics given the limited kinematic input set.
Within the same encoder configuration, these rows therefore mark the upper bound that the learned retargeting approaches without the run-time retargeting cost.

However, the policy trained on raw trajectories outperforms its retargeted counterpart on out-of-distribution motions.
As illustrated in Figure~\ref{fig:res:ood}, when mimicking a cantering gait excluded from training on the ANYmal D, the policy given retargeted inputs closely tracks the reference motion but eventually fails critically by falling.
In contrast, the policy trained on raw data internally reinterprets the motion as a stable trot, avoiding critical failure and demonstrating better generalization.
The higher success rates for the policies trained with the raw input sets in Table~\ref{tab:h1_mdme_abl_overall} may be attributed to this behavior.
Out-of-distribution generalization is further demonstrated using procedurally generated input trajectories, which are successfully used to generate feasible motions by our policy (shown in supplementary video).
Appendix~\ref{app:ssec:body_size} analyzes how estimated participant body size affects the policies trained with both the raw and retargeted input sets.

\begin{figure*}
    \centering
    \includegraphics[width=\linewidth]{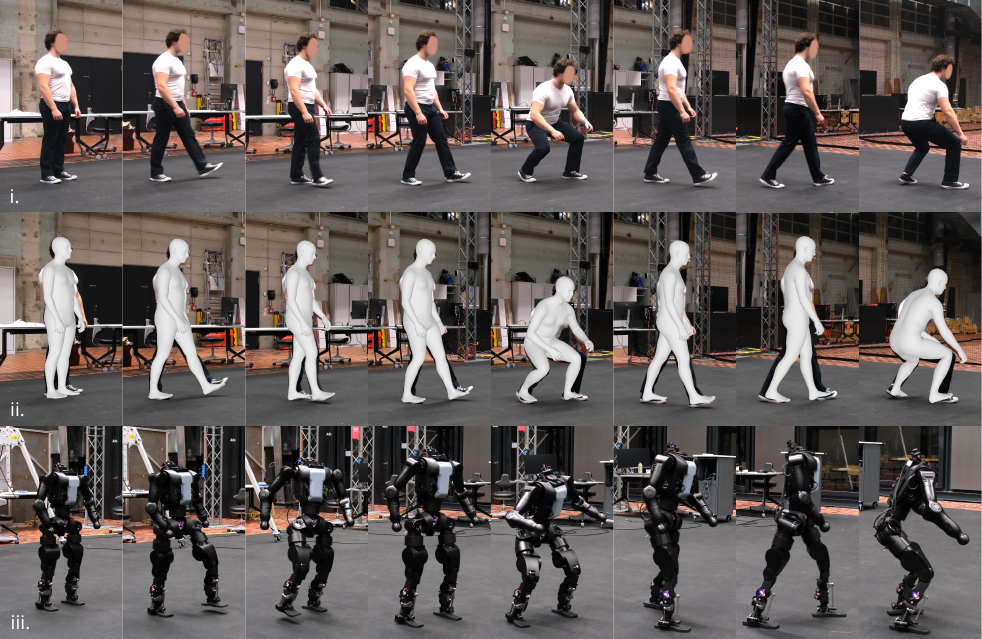}
    \caption{Zero-shot deployment of whole-body human mimicry on Fourier N1 platform: \textbf{i.} Reference motion from human actor. \textbf{ii.} Rendering of joint inputs on SMPL model over video. \textbf{iii.} Reference motion played on robot given input set described by Equation~\ref{eq:human_input}.}
    \label{fig:res:n1_sim2real}
\end{figure*}

\begin{table}[t]
  \centering
  \caption{Encoder-contribution attribution from evaluation-time permutation interventions on one trained MDME policy per platform (\texttt{db2} basis). At each step, the indicated encoder's output is shuffled across $1000$ parallel environments, destroying its information while preserving its statistics. Cells are means over five evaluation seeds of the single run.}
  \label{tab:attribution}
  \resizebox{\columnwidth}{!}{%
  \setlength{\tabcolsep}{3pt}
  \begin{tabular}{llcccccc}
    \toprule
    & Intervention & Jnt.\ Pos.$\downarrow$ & Lin.\ Vel.$\downarrow$ & Ang.\ Vel.$\downarrow$ & Orient.$\downarrow$ & Height$\downarrow$ & Succ.$\uparrow$ \\
    & & [rad] & [m/s] & [rad/s] & & [m] & \\
    \midrule
    \multirow{7}{*}{\rotatebox{90}{Humanoid}}
      & None (clean)            & 0.2035 & 0.3135 & 0.6056 & 0.1467 & 0.0578 & 0.9949 \\
      & $\mathbf{z}^v$ permuted & 0.2225 & 0.3386 & 0.6701 & 0.1542 & 0.0650 & 0.9763 \\
      & $\mathbf{z}^w$ permuted & 0.2036 & 0.3125 & 0.6082 & 0.1462 & 0.0576 & 0.9953 \\
      & Both permuted           & 0.2224 & 0.3377 & 0.6728 & 0.1533 & 0.0648 & 0.9750 \\
      \cmidrule{2-8}
      & $\phi_w$                &  0.0000 &  0.0010 & -0.0027 &  0.0007 &  0.0002 &  0.0005 \\
      & $\phi_v$                & -0.0189 & -0.0252 & -0.0646 & -0.0073 & -0.0072 &  0.0195 \\
    \midrule
    \multirow{7}{*}{\rotatebox{90}{Quadruped}}
      & None (clean)            & 0.1047 & 0.2640 & 0.6279 & 0.7950 & 0.0382 & 0.8954 \\
      & $\mathbf{z}^v$ permuted & 0.1087 & 0.2923 & 0.7160 & 0.7945 & 0.0387 & 0.9845 \\
      & $\mathbf{z}^w$ permuted & 0.1466 & 0.4129 & 0.7049 & 0.7929 & 0.0401 & 0.9486 \\
      & Both permuted           & 0.1526 & 0.4825 & 0.7949 & 0.7903 & 0.0397 & 0.9946 \\
      \cmidrule{2-8}
      & $\phi_w$                & -0.0429 & -0.1696 & -0.0780 &  0.0032 & -0.0015 & -0.0317 \\
      & $\phi_v$                & -0.0050 & -0.0490 & -0.0891 &  0.0016 & -0.0001 & -0.0676 \\
    \bottomrule
  \end{tabular}}
\end{table}

\subsection{Encoder-Contribution Attribution}
\label{ssec:attribution}

To evaluate MDME contributions within the trained policy, we shuffle one encoder's output across $1000$ parallel environments at every control step.
As the environments play different motions at different phases, the shuffled latent remains a valid sample from the encoder's output distribution, so the intervention severs the link to the current reference while preserving the input statistics.

Evaluating all four combinations of intact and permuted encoders forms a two-by-two factorial (Table~\ref{tab:attribution}).
Writing $e(S)$ for a metric evaluated with the set $S \subseteq \{w, v\}$ of encoders left intact, we report the Shapley attribution 
\begin{equation*}
    \phi_w = \tfrac{1}{2}\big[\big(e(\{w,v\}) - e(\{v\})\big) + \big(e(\{w\}) - e(\varnothing)\big)\big],
\end{equation*}
with $\phi_v$ defined analogously.
Each cell is evaluated over $10000$ steps on the held-out test split and averaged over five evaluation seeds; as all cells share this single checkpoint, the clean values differ from the pooled five-seed means of Table~\ref{tab:h1_mdme_abl_overall}.
The results suggest that the two platforms rely on opposite encoders.

On the humanoid, the four cells collapse to two levels set by whether $\mathbf{z}^v$ is intact: permuting $\mathbf{z}^w$ leaves every metric at its clean value ($0.2036$ against $0.2035$~rad joint position), while permuting $\mathbf{z}^v$ reproduces the fully deprived policy ($0.2225$ against $0.2224$~rad).
The trained humanoid policy therefore primarily relies on the unstructured encoder, although the wavelet encoder supports tracking when trained alone (Table~\ref{tab:h1_mdme_abl_overall}).
On the quadruped, the pattern reverses: the wavelet encoder provides most of the tracking information ($\phi_w = -0.0429$~rad joint position and $-0.1696$~m/s linear velocity, against $\phi_v = -0.0050$~rad and $-0.0490$~m/s).

Notably, this reliance does not follow from standalone performance.
The unstructured encoder performs better alone on the quadruped (Figure~\ref{fig:method_comparison}, Table~\ref{tab:h1_mdme_abl_overall}), but the jointly trained policy learns to rely on the wavelet encoder.
Joint training resolves the overlap by specializing to one encoder, and the preferred encoder matches the structure of the reference. The information-sparse, rhythmic quadruped reference suits a spectral representation, while the dense humanoid pose reference requires the higher-capacity unstructured embedding.

The entropy ablation of Section~\ref{res:ssec:ablations} supports this reading from the opposite direction.
Enriching the structured encoder by forwarding the raw DWT coefficients in place of the per-subband entropies improves quadruped tracking substantially, precisely where the attribution locates the policy's reliance on $\mathbf{z}^w$.
The same enrichment leaves the humanoid largely unchanged, consistent with a policy that draws its tracking information from $\mathbf{z}^v$ instead.
Improving the wavelet representation therefore pays off in proportion to how much the policy relies on it, a further indication that the wavelet encoder carries the quadruped task while contributing less to the humanoid one.

\section{Deployment Validation}
\label{res:ssec:hardware}
MDME policies for human and dog mimicry are deployed in a zero-shot manner on the Fourier N1 and ANYmal D platforms and successfully reproduce tested references in real-time.
On both platforms, the policy runs at $50$~Hz over a $500$~Hz joint-level PD controller (Table~\ref{tab:pd_gains}).

To imitate human motion, we extract SMPL joint information from RGB video \citep{shen2024gvhmr} and input the frame information to the trained policy.
Figure~\ref{fig:res:n1_sim2real} shows the successful recreation of new motions from demonstration without retargeting.
A qualitative analysis of the training dynamics and style-dependent performance across the motion datasets and an analysis of how participant body size affects imitation accuracy are provided in Appendix~\ref{app:sec:style_analysis}.

Additional experiments validate out-of-distribution performance of animal mimicry, given the limited motion styles used during training.
The policy is first tasked to achieve cross-platform imitation by transferring motions from a Unitree Go2 robot performing an out-of-distribution high-stepping gait to the ANYmal D.
Despite the entirely new motion style, the MDME architecture enables successful demonstration of this motion via degradation to a known, stable motion style.

To test motions from non-expert actors, we task the policy with reproducing motions from two human actors' legs.
After applying coordinate transformations to approximate a dog's skeletal structure, the policy successfully tracks the foot motions and body positions for basic walking, recreating both pacing and trotting gaits from human input in real-time.

\section{Conclusion}
\label{sec:conclusion}
This work introduces Multi-Domain Motion Embedding (MDME), a novel representation framework for real-time mimicry of human and animal motions on legged robots.
By fusing two complementary encoders, a discrete wavelet transform encoder for periodic structure and a probabilistic encoder for unstructured variation, MDME produces a stronger embedding than approaches using either alone.
This is demonstrated through improved tracking of human and animal motion data on humanoid and quadruped robotic platforms over our benchmarks.

We further use the improved motion representation of the MDME to track motions without pre-processed retargeting.
Compared to methods that use retargeted joint-level policy inputs, the proposed structure enables greater robustness over out-of-distribution motions when trained on raw motion data from the actor's morphology.
This also simplifies deployment, as a handcrafted real-time retargeting pipeline is not necessary.
We demonstrate this through zero-shot deployment on both humanoid and quadruped hardware.

Despite its success, our method is not without limitations and trade-offs.
The retargeting-free pipeline prioritizes real-time mimicry capability and robustness to motions outside of the distribution set that may appear, which may result in less accurate in-distribution results compared to methods using online retargeting to provide the policy input.
Difficult dynamic out-of-distribution motions are not perfectly tracked.
The policy instead degrades them to similar, dynamically feasible motions.
Additionally, it requires a significant degree of reward tuning from platform to platform to ensure that the desired behaviors are learned, and benefits from structural tuning for the wavelet encoder depending on the platform's motion features.

Future work may extend the versatility of the proposed architecture by investigating more diverse combinations of structured and unstructured embeddings within the MDME framework.
In particular, our attribution analysis shows that a jointly trained policy learns to rely on the encoder best suited to each platform's reference motions rather than drawing on both equally.
Selecting or designing encoders that provide distinct and synergistic motion cues may therefore yield more expressive and reliable embeddings.
Further independent development of structured motion representations that capture motion patterns without imposing overly rigid constraints could also enhance in-distribution policy performance and reduce the amount of reward tuning required.
Additional improvements may also be achieved by incorporating environmental compliance and increasing robustness to challenging terrains to advance overall humanoid control performance.

Ultimately, MDME and the proposed retargeting-free mimicry architecture provide a strong foundation for future research in robotic locomotion, real-time teleoperation, and other applications that require a concrete understanding of emerging patterns from motion.

\begin{acks}
The authors thank the members of the Robotic Systems Lab for their support and insights throughout the development of this work.
We are especially grateful to Chong Zhang and Shengzhi Wang for assistance with testing, and to Junzhe He, Elena Krasnova, and Vincent Griffo for their help in preparing the robot for experiments.
\end{acks}

\subsection*{\normalsize\sagesf\bfseries Author Contributions}\begin{refsize}\noindent
We confirm the contributions of this work's authors as follows:
conceptualization and manuscript review: M. Heyrman, C. Li, V. Klemm, D. Kang, S. Coros, and M. Hutter;
data curation and software: M. Heyrman, D. Kang;
formal analysis, validation, visualization, and manuscript preparation: M. Heyrman;
investigation and methodology: M. Heyrman, C. Li, V. Klemm, D. Kang;
supervision: S. Coros, M. Hutter;
funding acquisition: M. Hutter;
\end{refsize}

\begin{dci}
The authors declare that there are no potential conflicts of interest with respect to the research, authorship, and/or publication of this article.
\end{dci}

\begin{funding}
This project was supported in part by ETH AI Center, Fourier Intelligence, and ABB (via ETH Zurich Foundation) as part of the ETH RobotX research. 
\end{funding}

\bibliographystyle{SageH}
\bibliography{references}

\appendix
\section{Architecture Details}
\subsection{Structured Wavelet Encoder Implementation}
\label{app:ssec:dwt_implementation}

\begin{figure*}
    \centering
    \includegraphics[width=0.95\linewidth]{rebuttal/reb_graphics/fig_wavelet_bars_agg.pdf}
    \caption{Mother-wavelet ablation on both platforms. Aggregate tracking score of Equation~\ref{exp:eq:agg_perf} on the held-out test motions (higher is better), with error bars showing the standard deviation across training seeds; only the DWT basis changes between runs. The near-symmetric \texttt{sym2} performs best on the humanoid, and the symmetric \texttt{haar} and \texttt{rbio1.1} lead on the quadruped. We adopt \texttt{db2} as the standardized default for its balanced performance on both morphologies and its compact support shared with the per-platform winners.}
    \label{fig:mother_wavelet_ablation}
\end{figure*}

This section specifies the structured wavelet encoder $\mathbf{Enc}_w$ of Section~\ref{met:ssec:periodic} at the level of concrete tensor shapes, complementing the notation used in the main text.
The two-dimensional DWT is taken over the channel-time plane of the history buffer, so that the transform jointly resolves temporal and cross-channel structure, and the discrete transform is preferred over the continuous wavelet transform because it admits an exact, non-redundant, and differentiable fast filter-bank implementation.
Let $H$ denote the history-buffer length and $C = n_g$ the number of state channels per frame, so that the goal-state history $\mathbf{s}^g_t$ is arranged as a matrix $\mathbf{X} \in \mathbb{R}^{C \times H}$ (the batch dimension is omitted throughout).

Following the phase-extraction front end of periodic autoencoders \citep{PAE, FLD}, $\mathbf{X}$ is first passed through three one-dimensional convolution layers applied along the temporal axis, which reduce the channel dimension to $C_z$ latent channels while preserving the temporal length $H$.
The channel progression is $C \rightarrow C \rightarrow 2C_z \rightarrow C_z$, each layer using a kernel of width $H$, unit stride, and zero-padding of $\lfloor (H-1)/2 \rfloor$; the first two layers are followed by batch normalization and an ELU activation.
This yields the preprocessed embedding $\mathbf{Y} \in \mathbb{R}^{C_z \times H}$, where $C_z = 15$ for the humanoid and $C_z = 25$ for the quadruped (Table~\ref{exp:table:hyperparams}).
We note that $\mathbf{Y}$ has a fixed channel count $C_z$ and is the single quantity passed to the DWT; the layer index $l$ used in Section~\ref{met:ssec:periodic} refers only to these intermediate convolutions and does not appear in the transform output.

The preprocessed embedding $\mathbf{Y}$ is decomposed by a separable two-dimensional DWT over the channel-time plane across $J$ levels, using the selected mother wavelet and zero-padding at the boundaries, implemented with the differentiable filter-bank of \citet{pytorch_wavelet}.
Denoting the level-$l$ low-pass (approximation) and high-pass (detail) filters as in Section~\ref{met:ssec:periodic}, level $l$ produces one approximation subband $LL_l$ and three detail subbands $LH_l$, $HL_l$, $HH_l$, and the approximation is recursively decomposed so that the full output is
\[
\mathbf{W}^J_t = \{\,LL_J\,\} \cup \{\,LH_l, HL_l, HH_l\,\}_{l=1}^{J},
\]
a set of $1 + 3J$ subbands.
Each level subsamples both axes by two, so the number of resolvable levels grows with the temporal length $H$; the quadruped, with the longer buffer ($H=25$), uses $J=4$ levels, while the humanoid ($H=15$) uses $J=2$.

To obtain a fixed-size, low-dimensional encoding, each subband is summarized by a single wavelet-energy entropy \citep{wavelet-chinese-sl}.
For a subband with coefficients $\{w_i\}$, the normalized energy distribution and its Shannon entropy are
\[
p_i = \frac{w_i^2}{\sum_k w_k^2}, \qquad S = -\sum_i p_i \log_2 p_i,
\]
where a low entropy indicates energy concentrated in few coefficients (a persistent, structured oscillation) and a high entropy indicates energy spread across many coefficients (a transient, disordered event).
Our implementation also supports the Tsallis-$q$ generalization $S_q = \tfrac{1}{q-1}\big(1 - \sum_i p_i^{\,q}\big)$ \citep{tsallis-entropy}, which recovers the Shannon entropy as $q \to 1$; all reported results use $q = 1$.
Applying this descriptor to the $1 + 3J$ subbands yields the structured wavelet encoding $\mathbf{z}^w_t = \mathbf{S}_t \in \mathbb{R}^{1 + 3J}$: $7$ values for the humanoid and $13$ for the quadruped.
Here $\mathbf{S}_t$ denotes the vector of per-subband entropies; the scalar expression in Section~\ref{met:ssec:periodic} defines the entry of $\mathbf{S}_t$ computed for a single subband.

\begin{figure}
    \centering
    \includegraphics[width=\linewidth]{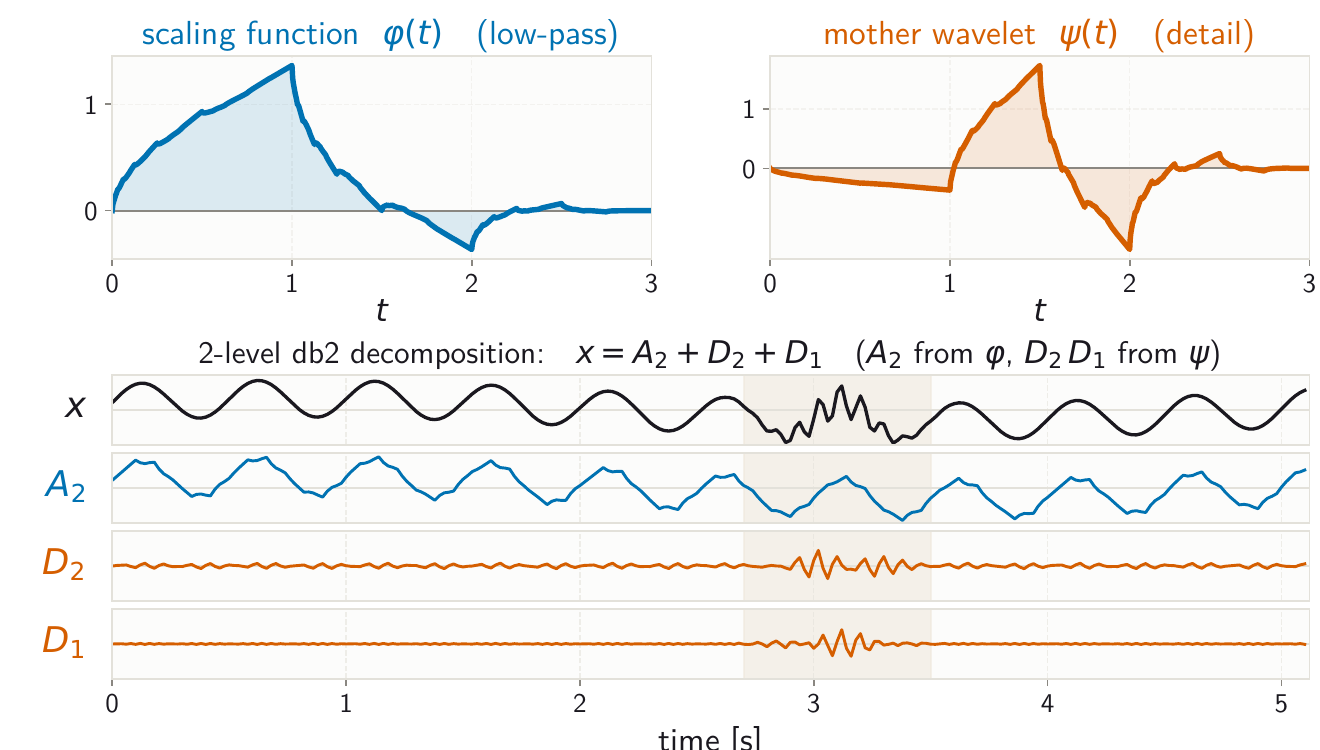}
    \caption{Top: the two \texttt{db2} basis functions, the scaling function $\varphi(t)$ used to generate the low-pass approximation band and the mother wavelet $\psi(t)$. Bottom: a 2-level \texttt{db2} transform of an illustrative signal. It decomposes as $x = A_2 + D_2 + D_1$; the approximation $A_2$ from $\varphi$ and detail bands $D_2$, $D_1$ from $\psi$.}
    \label{fig:app:db2_mother_wavelet}
\end{figure}
\begin{figure*}
    \centering
    \includegraphics[width=\linewidth]{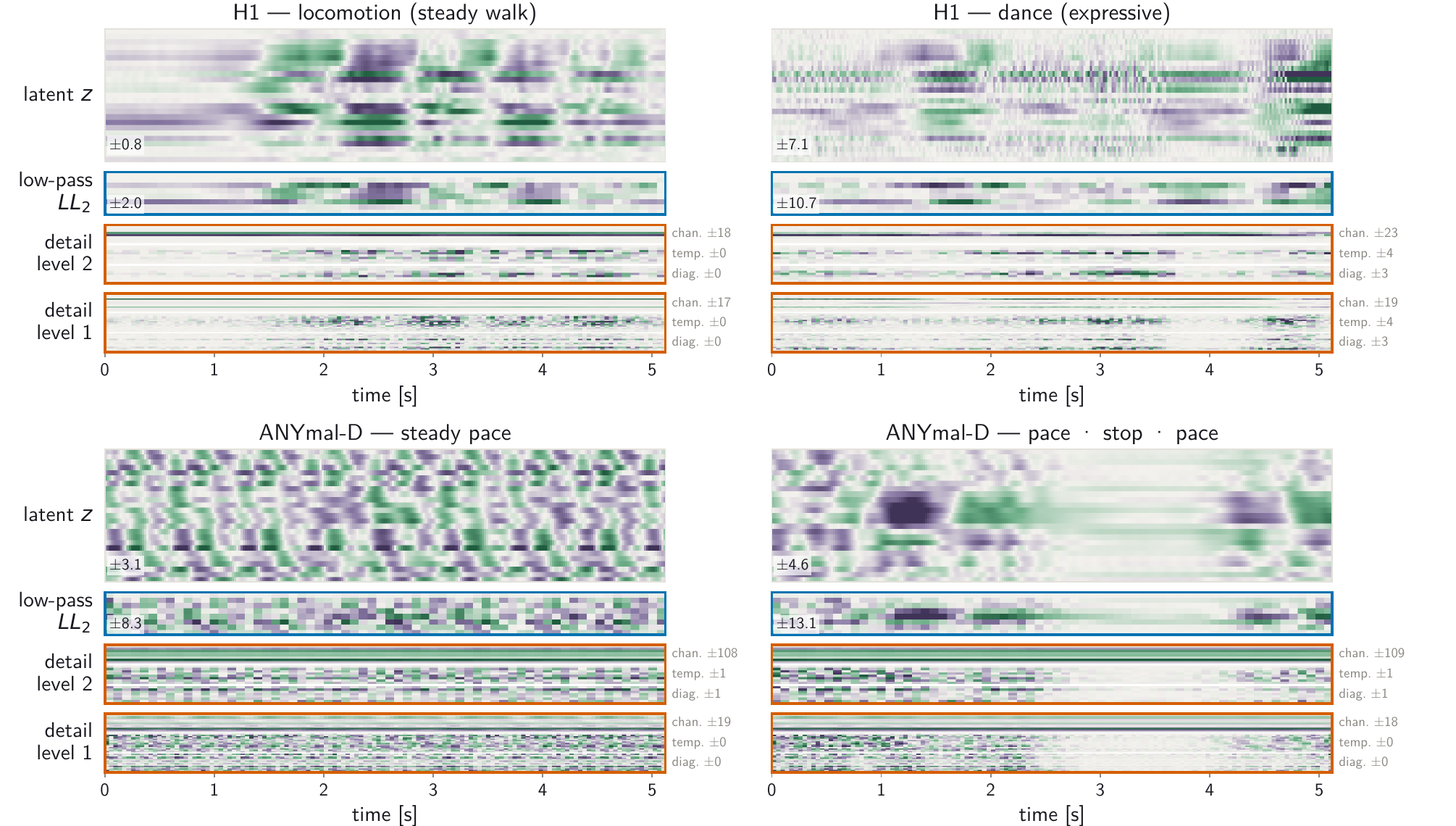}
    \caption{Wavelet decomposition using baseline \texttt{db2} mother wavelet for selected motions on humanoid and quadruped platform.}
    \label{fig:app:subbands}
\end{figure*}

\subsection{Mother Wavelet Choice}
\label{ssec:mother_wavelet_choice}

We select the mother wavelet through an ablation that varies only the DWT basis and holds the rest of MDME fixed, evaluated under deployment input noise on both platforms (Figure~\ref{fig:mother_wavelet_ablation}).
The sweep spans the Daubechies (\texttt{db2}, \texttt{db4}), Symlet (\texttt{sym2}), Coiflet (\texttt{coif1}--\texttt{coif3}), Haar, and biorthogonal (\texttt{bior1.1}, \texttt{rbio1.1}) families.

These families span the properties named in Section~\ref{met:ssec:periodic}.
A wavelet $\psi$ with $N$ vanishing moments satisfies $\int t^k \psi(t)\,dt = 0$ for $k < N$, so it ignores polynomial trends of degree below $N$; a larger $N$ captures smoother structure but requires a longer support, which for \texttt{db}$N$ takes the minimal value $2N-1$.
The Daubechies members therefore trade localization against smoothness directly: \texttt{haar} ($=\texttt{db1}$: $N{=}1$, support $1$), \texttt{db2} ($N{=}2$, support $3$), \texttt{db4} ($N{=}4$, support $7$).
\texttt{sym}$N$ keeps the same $N$ and support but is the least-asymmetric wavelet of that order, giving near-linear phase, and the biorthogonal pairs gain exact symmetry by relaxing orthogonality.

No wavelet performs best on both platforms.
On the humanoid, the near-symmetric \texttt{sym2} achieves the highest aggregate score; on the quadruped, \texttt{haar} and \texttt{rbio1.1} lead.
Each of these degrades markedly on the other morphology, while \texttt{coif3} and \texttt{db4} perform close to \texttt{db2} on both platforms.
The biorthogonal pair shows the strongest platform dependence: \texttt{bior1.1} and \texttt{rbio1.1} perform well on the quadruped but fall far behind every other basis on the humanoid.

This ordering follows from the wavelet properties.
On each platform, the best basis is symmetric or near-symmetric (\texttt{sym2} on the humanoid, \texttt{haar} and \texttt{rbio1.1} on the quadruped), following the symmetry of the training: the humanoid is trained with left--right symmetry data augmentation (Section~\ref{exp:ssec:human}) and the quadruped with reflection-based augmentation of the dataset, consistent with symmetric bases representing motion from a symmetric training distribution more compactly.

One trend is that smoothness separates the platforms.
The discontinuous \texttt{haar} suits the strongly periodic quadruped gaits but not the smoother humanoid motion, where the second vanishing moment of \texttt{sym2} and \texttt{db2} helps.
The same argument extends to the biorthogonal pair, where \texttt{bior1.1} and \texttt{rbio1.1} implement piecewise-constant filter pairs with a single vanishing moment, differing from \texttt{haar} mainly in the pairing of the analysis and synthesis filters.
These match the quadruped's sparse foot-position reference but represent the humanoid's dense joint-orientation input inefficiently, which may explain why both perform well on the quadruped but struggle on the humanoid.

Due to its consistent performance, the \texttt{db2} mother wavelet is used as a basis for explanations and ablations unless otherwise stated, shown in Figure~\ref{fig:app:db2_mother_wavelet}.
The decomposition of representative humanoid and quadruped motions under this basis is shown in Figure~\ref{fig:app:subbands}.
For future applications, we therefore recommend \texttt{db2} as a standard default, with a sweep over the compact low-order mother wavelets (\texttt{haar}, \texttt{db2}, \texttt{sym2}) to fine-tune performance for a particular platform.

\subsection{Comparison to Prior Motion Encoders}
\label{app:ssec:encoder_comparison}

This section formalizes the differences between the structured wavelet encoder and the frequency-domain encoders compared in Section~\ref{ssec:res:method_comp}.
All encoders operate on a latent channel $\mathbf{y} \in \mathbb{R}^{H}$ of the preprocessed embedding (Appendix~\ref{app:ssec:dwt_implementation}) and differ in the basis onto which the channel is projected and in how the resulting coefficients are summarized for the policy.

The PAE \citep{PAE} projects each channel onto the global Fourier basis and collapses the spectrum to a single sinusoid,
\[
\hat{\mathbf{y}}(\tau) = a \sin\!\big(2\pi(f\tau - \phi)\big) + b,
\]
with the frequency $f$ taken as the power-weighted mean of the spectrum, the amplitude $a$ from the total spectral power, the offset $b$ from the zero-frequency component, and the phase $\phi$ from a learned layer.
The basis functions $e^{i 2\pi f \tau}$ have support over the entire window, so the parametrization assumes one globally periodic component per channel; the energy of a transient or non-stationary segment is spread across the spectrum and absorbed into the averaged parameters rather than localized in time.

FLD \citep{FLD} retains PAE's parametrization but treats it as the state of a learned latent dynamical system, propagating $(f, a, \phi, b)$ forward autoregressively so that the representation describes the evolution of the periodic parametrization rather than a single fixed window.
Since this propagation defines the representation, the FLD baseline retains its native history length rather than our buffer $H$.

The short-time Fourier transform (STFT) projects the channel onto sinusoids restricted to a fixed-length analysis window,
localizing frequency content in time only up to that window length.
The resolution trade-off is therefore fixed once, uniformly over all frequencies, whereas the dyadic scaling of the DWT shortens the analysis support at fine scales and lengthens it at coarse scales, matching temporal resolution to frequency.

The Z-Transform (ZT) encoder projects the window onto damped complex exponentials with learnable poles.
A learned complex projection first forms modal coordinates $a_k(\tau)$ from the latent channels, and each mode is summarized by
\[
A_k = \sum_{\tau} a_k(\tau)\,\lambda_k^{\tau}, \qquad \lambda_k = r_k e^{i\theta_k},
\]
the window's Z-Transform evaluated at learnable poles inside the unit disk, with the time index normalized to $[0,1]$, the radius bounded in $r_k \in (0.5, 0.999)$ for stable gradients, and the normalized real and imaginary parts forwarded to the policy.

The DFT is the special case with the poles fixed on the unit circle, so each ZT mode can decay and represent a transient rather than a sustained oscillation.
The construction is inspired by Koopman-operator analyses, but no linear-dynamics or reconstruction constraint ties the poles to system eigenvalues; like every other encoder, the poles and projections are trained through the policy gradient alone.
Unlike the DFT and STFT variants, which retain the per-subband entropy summary, the ZT encoder forwards its modal coefficients directly, so this baseline varies the basis and the summary jointly.
The modal description remains global over the window and presumes an approximately stationary generative process, which non-stationary motion violates.

\begin{table}[t]
  \centering
  \caption{Policy parameter count and inference latency by encoder ablation on both platforms (deployed actor policy; batch-1 latency timed on an idle GPU). \textit{Deployed} is the total policy parameter count, split into the \textit{Encoder} and \textit{Policy MLP} counts. The MDME reference uses the \texttt{db2} basis. MDME (No ent.) forwards the raw DWT coefficients to the decoder in place of the per-subband entropies.}
  \label{tab:policy_cost}
  \resizebox{\columnwidth}{!}{%
  \setlength{\tabcolsep}{4pt}
  \begin{tabular}{llrrrr}
    \toprule
    Platform & Method & Deployed & Encoder & Policy MLP & Lat.@1 [ms]$\downarrow$ \\
    \midrule
    \multirow{9}{*}{\rotatebox{90}{Humanoid (H1)}}
      & MDME (Ours) & 571\,484 & 309\,577 & 261\,907 & 0.75 \\
      & MDME (No ent.) & 660\,060 & 309\,577 & 350\,483 & 0.53 \\
      & VAE & 611\,731 & 320\,640 & 291\,091 & 0.13 \\
      & PAE & 562\,601 & 328\,854 & 233\,747 & 0.48 \\
      & FLD & 255\,691 & 13\,752 & 241\,939 & 0.52 \\
      & DET & 533\,651 & 275\,328 & 258\,323 & 0.16 \\
      & MLP & 563\,659 & --- & 563\,659 & 0.19 \\
      & DFT & 561\,078 & 300\,707 & 260\,371 & 0.63 \\
      & STFT & 562\,614 & 300\,707 & 261\,907 & 0.64 \\
      & ZT & 574\,930 & 301\,247 & 273\,683 & 0.46 \\
    \midrule
    \multirow{9}{*}{\rotatebox{90}{Quadruped (ANYmal D)}}
      & MDME (Ours) & 625\,405 & 404\,849 & 220\,556 & 0.96 \\
      & MDME (No ent.) & 1\,089\,277 & 404\,849 & 684\,428 & 0.61 \\
      & VAE & 560\,876 & 346\,976 & 213\,900 & 0.14 \\
      & PAE & 616\,208 & 410\,500 & 205\,708 & 0.51 \\
      & FLD & 301\,507 & 52\,791 & 248\,716 & 1.00 \\
      & DET & 559\,820 & 345\,920 & 213\,900 & 0.14 \\
      & MLP & 653\,241 & --- & 653\,241 & 0.06 \\
      & DFT & 625\,917 & 404\,849 & 221\,068 & 0.81 \\
      & STFT & 625\,405 & 404\,849 & 220\,556 & 0.82 \\
      & ZT & 645\,749 & 406\,249 & 239\,500 & 0.46 \\
    \bottomrule
  \end{tabular}%
  }
\end{table}

\subsection{Computational Cost}
\label{app:ssec:compute_cost}

Table~\ref{tab:policy_cost} reports the parameter count and batch-1 GPU latency of the deployed policy for each encoder ablation on both platforms.
The learned methods have comparable parameter counts, confirming the parameter parity used in the method comparison, and every configuration runs far within the $20$~ms budget of the $50$~Hz control loop (MDME: $0.75$~ms humanoid, $0.96$~ms quadruped), so the encoder choice creates no run-time bottleneck.


\section{Training Details}
\label{app:sec:training}

The full implementation, including all reward definitions and training configurations, will be released publicly upon publication.

\begin{table*}[t]
  \centering
  \caption{Training-scheme ablation on the H1 humanoid: end-to-end (jointly trained) encoders against their two-stage (frozen, reconstruction-pretrained) and ELBO (jointly trained with an auxiliary decoder reconstructing the retargeted reference) counterparts, reported as the pooled per-motion mean~$\pm$~the standard deviation of the per-seed evaluation means (MAE for joint position and height; L2-norm for velocity and orientation). The policies are trained without domain randomization and input noise to better capture reconstruction performance. MDME uses the standardized \texttt{db2} basis in all rows so the comparison isolates the training scheme. \textbf{Bold}: best training scheme per encoder and metric.}
  \label{tab:pretrain_ablation}
  \resizebox{\textwidth}{!}{%
  \begin{tabular}{llcccccc}
    \toprule
    Encoder & Training & Jnt.\ Pos.\ [rad]$\downarrow$ & Lin.\ Vel.\ [m/s]$\downarrow$ & Ang.\ Vel.\ [rad/s]$\downarrow$ & Orientation$\downarrow$ & Height [m]$\downarrow$ & Success$\uparrow$ \\
    \midrule
    \multirow{3}{*}{MDME (\texttt{db2})} & End-to-end & 0.2036$\pm$0.0007 & \textbf{0.3149$\pm$0.0018} & \textbf{0.7575$\pm$0.0034} & \textbf{0.1482$\pm$0.0011} & \textbf{0.0633$\pm$0.0001} & \textbf{0.9969} \\
     & Frozen PT & 0.2072$\pm$0.0066 & 0.3346$\pm$0.0054 & 0.8289$\pm$0.1071 & 0.1544$\pm$0.0024 & 0.0666$\pm$0.0004 & 0.9920 \\
     & ELBO & \textbf{0.2016}$\pm$0.0006 & 0.3211$\pm$0.0022 & 0.8183$\pm$0.0057 & 0.1534$\pm$0.0010 & 0.0641$\pm$0.0001 & 0.9882 \\
    \midrule
    \multirow{3}{*}{VAE} & End-to-end & 0.2076$\pm$0.0070 & 0.3230$\pm$0.0057 & 0.8135$\pm$0.1068 & 0.1493$\pm$0.0022 & 0.0629$\pm$0.0002 & \textbf{0.9971} \\
     & Frozen PT & 0.2195$\pm$0.0068 & \textbf{0.3203$\pm$0.0060} & 0.8251$\pm$0.1062 & 0.1617$\pm$0.0025 & 0.0640$\pm$0.0002 & 0.9918 \\
     & ELBO & \textbf{0.2018$\pm$0.0008} & 0.3266$\pm$0.0012 & \textbf{0.8112$\pm$0.0026} & \textbf{0.1472$\pm$0.0008} & \textbf{0.0596$\pm$0.0001} & 0.9958 \\
    \midrule
    \multirow{3}{*}{PAE} & End-to-end & \textbf{0.2282$\pm$0.0062} & \textbf{0.3269$\pm$0.0053} & \textbf{0.8357$\pm$0.1062} & 0.1618$\pm$0.0026 & 0.0621$\pm$0.0001 & \textbf{0.9980} \\
     & Frozen PT & 0.2660$\pm$0.0043 & 0.3667$\pm$0.0045 & 0.8736$\pm$0.1051 & 0.1633$\pm$0.0028 & 0.0635$\pm$0.0002 & 0.9914 \\
     & ELBO & 0.2283$\pm$0.0007 & 0.3337$\pm$0.0015 & 0.8370$\pm$0.0049 & \textbf{0.1486$\pm$0.0009} & \textbf{0.0618$\pm$0.0001} & 0.9931 \\
    \bottomrule
  \end{tabular}}
\end{table*}

\subsection{End-to-End vs. Two-Stage Encoder Training}
\label{app:ssec:stage_training}

Prior representation learning methods for motion \citep{VAE, PAE, FLD} decouple the encoder from the control task:
the encoder is first trained offline as an autoencoder that reconstructs the motion, then frozen, and only the policy is trained on the fixed latent representation.
MDME instead trains the encoder jointly with the policy, so that the encoder gradients flow from the PPO objective and the representation is shaped by the combined retargeting and mimicry task rather than by reconstruction alone.

To quantify the value of these task-driven gradients, we run a two-stage ablation on the humanoid that reproduces the prior-work paradigm for each encoder.
Following the pipeline of \citet{VMP} for the unstructured encoder and \citet{PAE} for the periodic encoder baselines, we (i) extract the reference-motion windows the encoder consumes, (ii) train the encoder as a plain autoencoder with a mean-squared reconstruction objective, and (iii) freeze the encoder and train only the actor and critic heads with PPO.
Each frozen variant reuses the exact architecture of its jointly trained counterpart, differing only in the training signal; the pretraining window is that of the original method ($61$ frames for the periodic and unstructured encoders, $15$ for the MDME encoder).

This comparison retains the per-component error metrics rather than the aggregate score of Section~\ref{exp:ssec:eval_method}:
the mean absolute error (MAE) for the joint positions and base height, the $\ell_2$-norm error for the base linear velocity, angular velocity, and orientation, and the survival success rate, the fraction of episodes that terminate without a fall.
All metrics are computed per motion and pooled over the test motions for each of five independently trained policies, and each error metric is reported as the pooled mean together with the standard deviation of these per-seed evaluation means; the survival success rate is reported as the pooled mean.

Table~\ref{tab:pretrain_ablation} compares the frozen, pretrained encoders against their end-to-end counterparts, using the standardized \texttt{db2} basis for MDME in all rows so the comparison isolates the training scheme.
End-to-end training lowers the joint position and angular velocity errors for every encoder, confirming that the representation benefits from task-driven gradients that a reconstruction objective does not provide.
The gap is largest for the periodic PAE encoder (joint position error $0.2282 \rightarrow 0.2660$, $+17\%$) and smallest for MDME ($0.2036 \rightarrow 0.2072$, $\sim\!2\%$), which degrades most gracefully and remains the strongest frozen variant.
This supports our central design choice: since MDME must reinterpret raw references as feasible actions on the robot, coupling the encoder to the control objective proves more effective than learning a task-agnostic motion manifold in a separate stage.

A third scheme keeps the joint training and adds the reconstruction signal as an auxiliary objective rather than a separate stage.
The ELBO variants of the MDME, VAE, and PAE encoders train a decoder to reconstruct the retargeted reference states from the latent embedding alongside PPO using a decoder with two hidden layers of shape $[256, 256]$ during training.
Reconstructing the retargeted reference, rather than the raw input window, ties the auxiliary signal to the same target the tracking rewards use.
The resulting results are highly encoder dependent, offering marginal benefits to the joint position reconstruction for the MDME case and higher benefits for the PAE case, outperforming the VAE end-to-end example.

This suggests that structured encoders see fewer benefits to being given an explicit reconstruction target compared to unstructured encoders.
This may potentially be due to the improved modeling of the inherent temporal structures in the input motion signal.
Based on these comparisons, we keep the purely task-driven training as the default.

\begin{table}[]
    \centering
    \small
    \begin{tabular}{lcc}
        \toprule
        Reward Term & Weight $w$ & Scale $\sigma$ \\
        \midrule
        Foot Tracking & 1.5 & 2.0 \\
        Joint Tracking & 2.0 & 5e-1 \\
        Velocity Tracking & 2.5 & 7.5e-1 \\
        Angular Velocity Tracking & 3.0 & 5e-1 \\
        Projected Gravity Tracking & 1.0 & 1e-2 \\
        Base Height Tracking & 1.5 & 1e-1 \\
        \midrule
        Joint Torques & -3e-5 & --- \\
        Joint Acceleration & -4e-7 & --- \\
        Joint Action Rate & -1.5e-2 & --- \\
        Undesired Contacts & -1.0 & --- \\
        Termination & -1e3 & --- \\
        \bottomrule
    \end{tabular}
    \caption{Quadruped training rewards and weights.}
    \label{exp:table:quad_rewards}
\end{table}
\begin{table}[]
    \centering
    \small
    \begin{tabular}{lcc}
        \toprule
        Reward Term & Weight $w$ & Scale $\sigma$ \\
        \midrule
        Upper Body Joint Tracking & 5.0 & 2.0 \\
        Lower Body Joint Tracking & 7.0 & 1.5 \\
        X-Velocity Tracking & 10.0 & 5e-2 \\
        Y-Velocity Tracking & 2.0 & 5e-2 \\
        Z-Velocity Tracking & 2.0 & 1e-1 \\
        Angular Velocity Tracking & 10.0 & 2.5e-1 \\
        Projected Gravity Tracking & 3.0 & 5e-2 \\
        Base Height Tracking & 2.0 & 2.5e-1 \\
        \midrule
        Feet Dist Tracking & -5e-1 & 2.5e-1 \\
        Feet Projected Gravity & -1e-2 & 1.0 \\
        Feet Air Time Reward & 5.0 & --- \\
        Feet Sliding & -5e-1 & --- \\
        Joint Torques & -1e-5 & --- \\
        Joint Acceleration & -2e-7 & --- \\
        Joint Action Rate & -1e-2 & --- \\
        Termination & -1e3 & --- \\
        \bottomrule
    \end{tabular}
    \caption{Humanoid training rewards and weights for Unitree H1 platform.}
    \label{exp:table:h1_rewards}
\end{table}
\begin{table}[]
    \centering
    \small
    \begin{tabular}{lcc}
        \toprule
        Reward Term & Weight $w$ & Scale $\sigma$ \\
        \midrule
        Upper Body Joint Tracking & 15.0 & 2.0 \\
        Lower Body Joint Tracking & 30.0 & 1.5 \\
        X-Velocity Tracking & 15.0 & 5e-1 \\
        Y-Velocity Tracking & 10.0 & 2.0 \\
        Z-Velocity Tracking & 2.0 & 5e-1 \\
        Angular Velocity Tracking & 10.0 & 2.0 \\
        Foot Height Tracking & 20.0 & 5e-4 \\
        Projected Gravity Tracking & 0.5 & 1e-1 \\
        Base Height Tracking & 0.5 & 2.5e-1 \\
        \midrule
        Feet Dist Tracking & -5e-1 & 2.5e-1 \\
        Feet Projected Gravity & -2.0 & 5e-1 \\
        Feet Air Time Reward & 15.0 & --- \\
        Feet Sliding & -5.0 & --- \\
        Joint Torques & -1e-3 & --- \\
        Joint Acceleration & -1e-5 & --- \\
        Joint Action Rate & -1e-4 & --- \\
        Joint Position Limits & -2.0 & --- \\
        Termination & -1e3 & --- \\
        \bottomrule
    \end{tabular}
    \caption{Humanoid training rewards and weights for Fourier N1 platform.}
    \label{exp:table:n1_rewards}
\end{table}

\subsection{Reward Definitions}
\label{app:ssec:rewards}

The per-platform weights $w$ and scales $\sigma$ of the reward terms described in Section~\ref{sec:experiments} are listed in Tables~\ref{exp:table:quad_rewards}--\ref{exp:table:n1_rewards}.
Each reference-tracking term applies a Gaussian kernel to the error $\mathbf{e} = \mathbf{s}^g_{ret} - \mathbf{s}^p$ between the robot's state and the corresponding component of the ideal retargeted reference,
\begin{equation}
\label{exp:eq:reward}
    r^{\text{track}} = w \, \exp\!\left(-\frac{\lVert \mathbf{e} \rVert}{\sigma}\right).
\end{equation}

The regularization penalties, in contrast, act directly on the robot state rather than through a kernel. The joint-torque, joint-acceleration, and action-rate terms penalize $\lVert \boldsymbol{\tau}_t \rVert^2$, $\lVert \ddot{\mathbf{q}}_t \rVert^2$, and $\lVert \mathbf{a}_t - \mathbf{a}_{t-1} \rVert^2$ respectively, and the joint-position-limit term (N1) penalizes joint targets driven past their soft limits.

The gait-shaping terms of the humanoid platforms reward or penalize discrete events. The feet-air-time reward encourages discrete stepping over foot shuffling by rewarding the swing duration of each foot while it stays below a $0.5$~m height threshold; the feet-sliding penalty discourages horizontal foot motion during stance; the inter-foot-distance penalty discourages an overly wide stance; and the feet-orientation term, computed from the feet projected gravity, keeps the soles level.

\section{Motion Style Analysis}
\label{app:sec:style_analysis}

\begin{figure*}
    \centering
    \includegraphics[width=\linewidth]{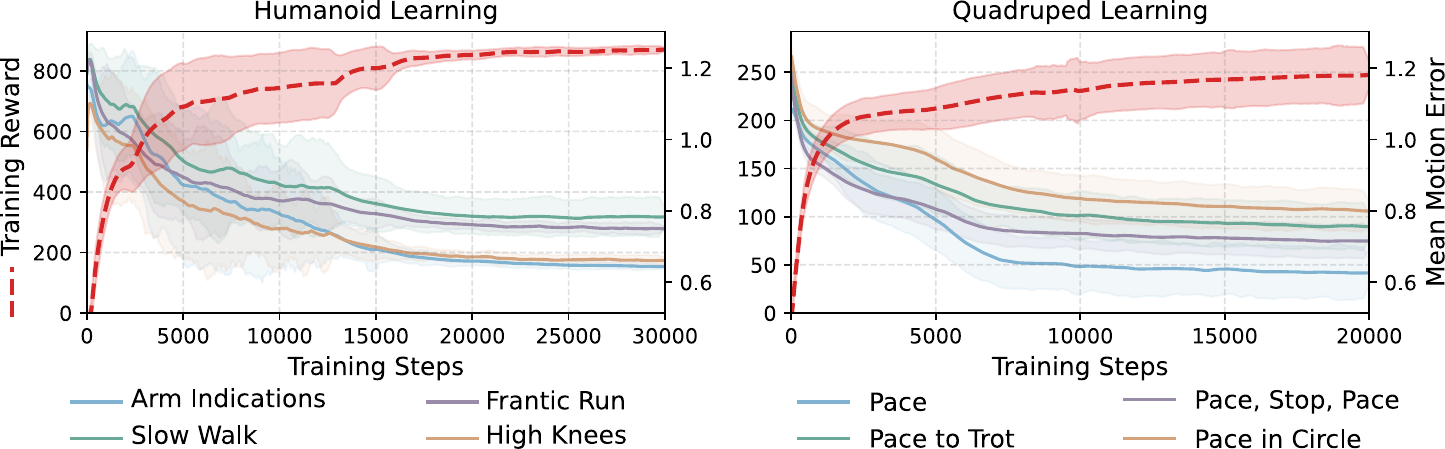}
    \caption{Learning curves of humanoid and quadruped mimicry training (plotted in red) and the performance for four selected reference motions of varying styles demonstrating learned performance over time.}
    \label{fig:res:learning_curves}
\end{figure*}

To better understand how MDME policies acquire and reproduce different motion styles, we examine their training dynamics and performance variation across the AMASS and MANN datasets.
This analysis is qualitative: we restrict the projection's interpretation to coarse trends, anchored to the measured per-motion tracking score of Section~\ref{exp:ssec:eval_method} rather than to the visual layout alone.
Figure~\ref{fig:res:learning_curves} presents learning curves for representative humanoid and quadruped motions.
These curves show that the policies not only achieve stable locomotion early in training but also continue to refine performance over time, adapting to both simple and complex behaviors under the concurrent training schedule shared by all methods.

\begin{figure}
    \centering
    \includegraphics[width=\linewidth]{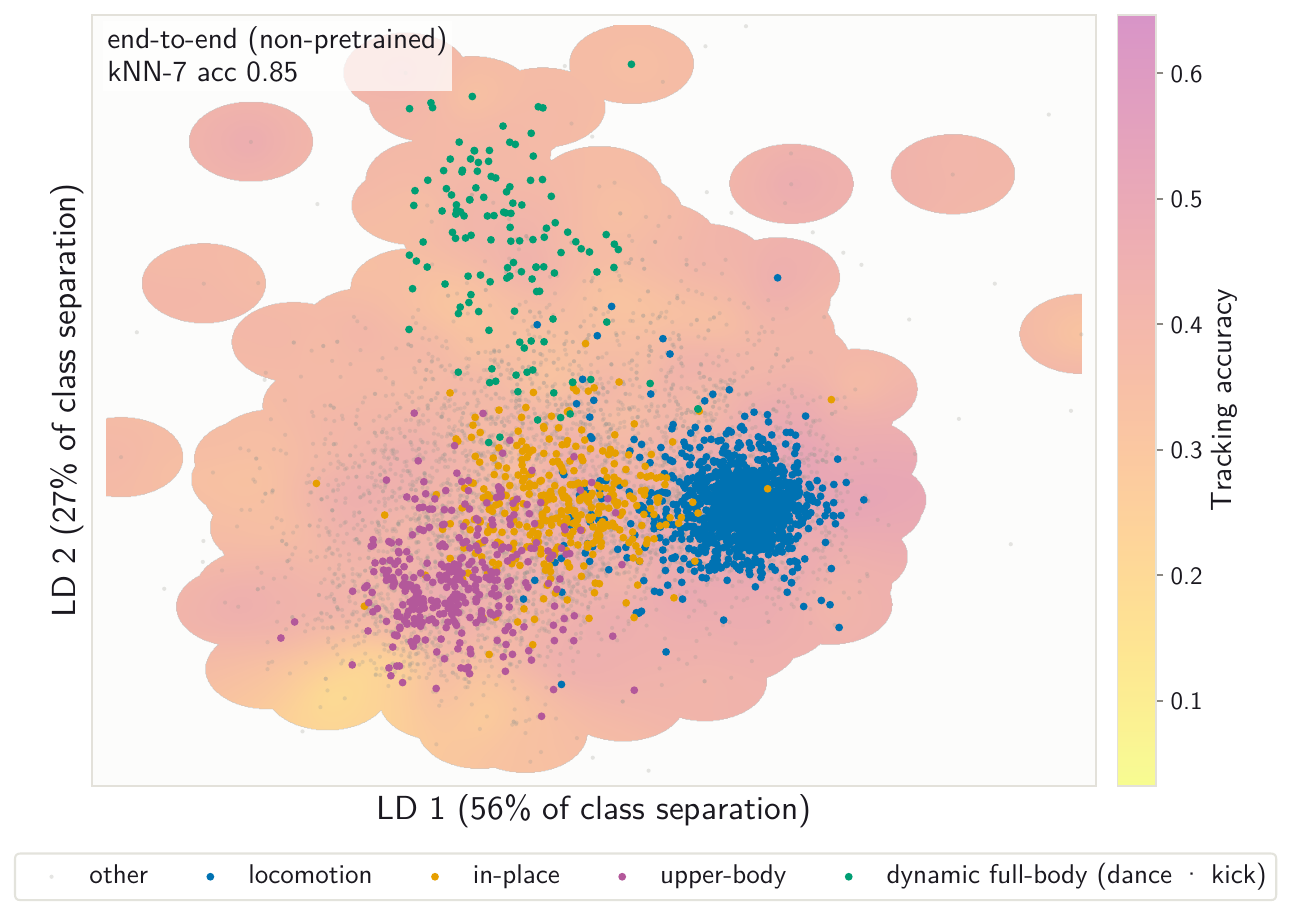}
    \caption{Unseen human test motions projected from the embeddings of the end-to-end encoder colored by style group: \textbf{locomotion} (walking, running, turning), \textbf{in-place} (standing, posing, jumping), \textbf{upper-body} (gesturing, throwing, punching), and \textbf{dynamic full-body} (dance, kicks). Motions matching no label are in gray. The underlay interpolates the per-motion tracking score of Section~\ref{exp:ssec:eval_method} between scored motions.}
    \label{fig:res:heatmap}
\end{figure}

We assess style-dependent performance over the labeled styles of the unseen human motions.
Style labels are assigned from filename keywords and merged into four groups, locomotion, in-place, upper-body, and dynamic full-body motions, with the grouping selected by cross-validated nearest-neighbor accuracy; motions matching no rule remain unlabeled.
Figure~\ref{fig:res:heatmap} projects the test motions from the embeddings of the end-to-end encoder over a field interpolating the per-motion tracking score of Section~\ref{exp:ssec:eval_method}.
Locomotion forms the most compact and separable cluster and lies over the most accurately tracked region, consistent with periodic gait producing the most distinctive wavelet signature.
In-place motions bridge locomotion and the upper-body cluster, with mid-range accuracy.
Upper-body motions track well despite their expressiveness, as balance is barely challenged.
The dynamic full-body group, dance sequences and kicks with whole-body balance transfer, lies farthest from locomotion, over the weakest region of the field, and is the hardest to track.

Further analysis of the learning curves for humanoid and quadruped imitation reveals how the policy learns various motion styles during training.
The humanoid motion mimicry learning curve in Figure~\ref{fig:res:learning_curves} shows that after an initial period ($\approx$3\,000 steps) of learning basic standing and locomotion, performance gradually improves before a significant reward jump after 10\,000 steps.
In agreement with Figure~\ref{fig:res:heatmap}, the policy motions with exaggerated movements (e.g., high-knee walks, arm gestures) more readily, with generally higher error for less stable dynamic motions or slow and in-place locomotion that requires more balance.

For dog mimicry, the policy rapidly achieves stable locomotion and initial mimicry within the first few thousand steps.
Motions with a single, stable gait (e.g., Pace) are quickly learned with better performance.
Around 2\,500 steps, a performance jump occurs for transitions (e.g., Pace to Trot), indicative of learning general gait mimicry.
More complex motions, such as pacing in a circle where the dog uses spinal flexibility to look opposite its turn, are learned more slowly.
The sparse input representation does not explicitly capture this torsion, forcing the policy to learn the relationship purely from foot positions.
Performance on this complex motion improves significantly around 10\,000 steps, eventually matching the performance of gait transitions.

\subsection{Participant Body Size}
\label{app:ssec:body_size}

The AMASS dataset uses motions from participants of widely varying body dimensions.
To measure how participant size relates to imitation performance, we estimate each test motion's participant height, weight, and limb proportions from the sequence's SMPL+H shape parameters and gender, using the canonical zero-pose mesh \citep{bittner2023camerahumankinematics, schneider2024musclestimelearningunderstand}.
The $1\,933$ H1 test motions ($1\,113$ male, $820$ female) span estimated heights of $1.75 \pm 0.09$~m.
Table~\ref{tab:body_size} reports the Spearman correlation between each body measure and the per-motion tracking errors of Section~\ref{exp:ssec:eval_method}.

These results show that the body size strongly predicts imitation accuracy.
Taller, longer-limbed participants are tracked with substantially lower error ($\rho = -0.56$ between height and joint position error, $-0.57$ for orientation), and between the shortest and tallest height deciles the orientation error falls from $0.315$ to $0.098$ and the joint position error from $0.308$ to $0.188$.
Stature and limb length dominate over mass, as BMI correlates far more weakly, and hip width is the lone reversal, consistent with a mismatch of proportion rather than scale.
The dependence survives controlling for motion style through partial correlations, and the overall success rate shows no comparable dependence, so the effect appears in tracking accuracy rather than in falls.
Gender differences follow the size distribution rather than gender itself.

To test whether this dependence is an artifact of conditioning on the raw SMPL reference, we repeat the analysis on the retargeted-input policy of Section~\ref{res:ssec:ablations}, which receives robot-space joint targets.
The correlation structure reproduces almost exactly (agreement $r = 0.90$ across all $63$ measure-metric pairs, $49$ significant in both), the orientation dependence is stronger still ($\rho = -0.63$), and the shortest-decile orientation error is identical ($0.315$).
Retargeted inputs attenuate only the joint position dependence ($-0.56$ to $-0.39$).

We therefore attribute the effect to retargeting human motion onto a fixed-size humanoid rather than to the input representation.
A participant near the H1's scale maps to more kinematically feasible robot-space trajectories, while a smaller participant's motion is warped further and drives the robot toward less stable postures.

\begin{table}[t]
  \centering
  \caption{Spearman correlation between estimated participant body measures and per-motion tracking error on the H1 test set for the retargeting-free MDME policy and the retargeted-input policy of Section~\ref{res:ssec:ablations}. Negative values mean larger participants are imitated more accurately.}
  \label{tab:body_size}
  \resizebox{\columnwidth}{!}{%
  \begin{tabular}{lcccccc}
    \toprule
    & \multicolumn{3}{c}{MDME (raw input)} & \multicolumn{3}{c}{Retargeted input} \\
    \cmidrule(lr){2-4}\cmidrule(lr){5-7}
    Body measure & Jnt.\ Pos. & Orientation & Lin.\ Vel. & Jnt.\ Pos. & Orientation & Lin.\ Vel. \\
    \midrule
    Height & $-0.558$ & $-0.567$ & $-0.279$ & $-0.393$ & $-0.634$ & $-0.245$ \\
    Leg length & $-0.502$ & $-0.529$ & $-0.242$ & $-0.365$ & $-0.617$ & $-0.219$ \\
    Arm length & $-0.506$ & $-0.486$ & $-0.271$ & $-0.379$ & $-0.594$ & $-0.246$ \\
    Torso length & $-0.501$ & $-0.480$ & $-0.266$ & $-0.357$ & $-0.525$ & $-0.228$ \\
    Shoulder width & $-0.433$ & $-0.442$ & $-0.231$ & $-0.295$ & $-0.492$ & $-0.213$ \\
    Weight & $-0.491$ & $-0.481$ & $-0.222$ & $-0.323$ & $-0.451$ & $-0.181$ \\
    BMI & $-0.299$ & $-0.285$ & $-0.115$ & $-0.171$ & $-0.187$ & $-0.082$ \\
    Hip width & $+0.121$ & $+0.142$ & $+0.105$ & $+0.100$ & $+0.229$ & $+0.117$ \\
    \bottomrule
  \end{tabular}}
\end{table}

\end{document}